\documentclass[lettersize,journal]{IEEEtran}
\usepackage{amsmath,amsfonts}
\usepackage{algorithmic}
\usepackage{algorithm}
\usepackage{array}
\usepackage[caption=false,font=normalsize,labelfont=sf,textfont=sf]{subfig}
\usepackage{textcomp}
\usepackage{stfloats}
\usepackage{url}
\usepackage{verbatim}
\usepackage{graphicx}
\usepackage{cite}
\usepackage{amssymb}
\usepackage{mathtools}
\usepackage{amsthm}

\usepackage{thm-restate}
\theoremstyle{plain}
\newtheorem{theorem}{Theorem}[section]

\theoremstyle{definition}

\theoremstyle{remark}
\newtheorem{remark}[theorem]{Remark}

\usepackage[capitalize,noabbrev]{cleveref}
\crefname{section}{Sec.}{Secs.}
\Crefname{section}{Section}{Sections}
\crefname{table}{Tab.}{Tabs.}
\Crefname{table}{Table}{Tables}
\crefname{figure}{Fig.}{Figs.}
\Crefname{figure}{Figure}{Figures}
\crefname{equation}{Eq.}{Eqs.}
\Crefname{equation}{Equation}{Equations}
\crefname{assumption}{Assumption}{Assumptions}
\Crefname{assumption}{Assumption}{Assumptions}

\usepackage{microtype}
\usepackage{booktabs}
\usepackage{multirow} 
\usepackage[table]{xcolor}

\begin{document}

\title{Diffusion Secant Alignment for Score-Based Density Ratio Estimation}

\newcommand{\SCUT}{%
	\textsuperscript{\ref{fn:scut}}%
}

\author{
	Wei Chen\textsuperscript{\rm 1},
	Shigui Li\textsuperscript{\rm 1},
	Jiacheng Li\textsuperscript{\rm 1},
	Jian Xu\textsuperscript{\rm 1}, 
	Zhiqi Lin\textsuperscript{\rm 1},  \\
	Junmei Yang\textsuperscript{\rm 1}, 
	Delu Zeng\textsuperscript{\rm 1}\thanks{Corresponding author.}, 
	John Paisley\textsuperscript{\rm 2}, 
	Qibin Zhao\textsuperscript{\rm 3}
	\thanks{%
			\textsuperscript{\rm 1}South China University of Technology, \textsuperscript{\rm 2}Columbia University, \textsuperscript{\rm 3}AIP, RIKEN.
		}
}

\markboth{}%
{Shell \MakeLowercase{\textit{et al.}}: A Sample Article Using IEEEtran.cls for IEEE Journals}


\maketitle

\begin{abstract}
Estimating density ratios has become increasingly important with the recent rise of score-based and diffusion-inspired methods. However, current tangent-based approaches rely on a high-variance learning objective, which leads to unstable training and costly numerical integration during inference. We propose \textit{Interval-annealed Secant Alignment Density Ratio Estimation (ISA-DRE)}, a score-based framework along diffusion interpolants that replaces the instantaneous tangent with its interval integral, the secant, as the learning target. We show theoretically that the secant is a provably lower variance and smoother target for neural approximation, and also a strictly more general representation that contains the tangent as the infinitesimal limit. To make secant learning feasible, we introduce the \textit{Secant Alignment Identity (SAI)} to enforce self consistency between secant and tangent representations, and \textit{Contraction Interval Annealing (CIA)} to ensure stable convergence.  Empirically, this stability-first formulation produces high efficiency and accuracy. ISA-DRE achieves comparable or superior results with fewer function evaluations, demonstrating robustness under large distribution discrepancies and effectively mitigating the density-chasm problem.
\end{abstract}


\section{Introduction}

\IEEEPARstart{E}{stimating} the density ratio, $r(\boldsymbol{x}) = p_1(\boldsymbol{x})/p_0(\boldsymbol{x})$, is fundamental in machine learning and neural computing, underpinning diverse systems such as density estimation  \cite{peerlings2022multivariate,letizia2025copula}, transfer learning \cite{zhang2022transfer}, neural topic modeling \cite{wang2024mutual}, multilabel causal feature selection \cite{ma2025mi} and multi-agent reinforcement learning \cite{ding2023multiagent,li2025robust}.
A conventional approach estimates $p_0$ and $p_1$ directly with a neural network, but becomes computationally fragile when the distributions exhibit significant discrepancies, which is a challenge known as the \textit{density-chasm problem} \cite{rhodes2020telescoping}.

Direct density ratio estimation (DRE) methods such as noise-contrastive estimation \cite{gutmann2010noise} and trimmed estimators \cite{liu2017trimmed} mitigate this issue but often suffer from high variance or require heavy hyperparameter tuning. TRE \cite{rhodes2020telescoping} alleviates density chasms by factorizing the global ratio into local ratios across intermediate distributions, yet demands training multiple models and remains sensitive to distribution shifts \cite{choi2022density}.
Score-based DRE frameworks take a different route. DRE-$\infty$ \cite{choi2022density} rewrites the log-ratio as a time integral $\log r(\boldsymbol{x}) = \int_0^1 \partial_t \log p_t(\boldsymbol{x}) \mathrm{d}t$, where $\partial_t \log p_t$ is the \emph{time score}.
This enables single-model training along a continuous path $p_t$, typically derived with a diffusive interpolant, and has inspired several variants, including diffusion-bridge interpolants \cite{chen2025dequantified,yu2025density}.
Despite these advances, all such methods share a fundamental bottleneck: \emph{they learn the instantaneous tangent $\partial_t \log p_t$, an intrinsically high-variance and noisy target}.
The resulting instability yields unstable gradients \cite{glorot2010understanding} and forces heavy numerical quadrature at inference time, not for integration per se, but to average out accumulated estimation noise \cite{zhu2022numerical}.

In this work, we address the root cause of instability by shifting the learning target from the instantaneous tangent to its interval integral. We define the \textbf{secant} function as $u(\boldsymbol{x},l, t)=\frac{1}{t-l}\int_{l}^{t}\partial_{\tau}\log p_{\tau}(\boldsymbol{x})\mathrm{d}{\tau}, l < t$, 
representing the average rate of log-density change over $[l, t]\subseteq [0,1]$. As shown in \cref{eq:tangent-secant-relationship}, the secant is the expectation of all tangents on the interval, and the log-ratio is simply recovered as  $\log r(\boldsymbol{x})=u(\boldsymbol{x},0,1)$. 

\begin{figure*}[!t]
	\centering
        \subfloat[Score estimation + integration.\label{fig:illustration-tangent}]{%
          \includegraphics[width=0.32\linewidth]{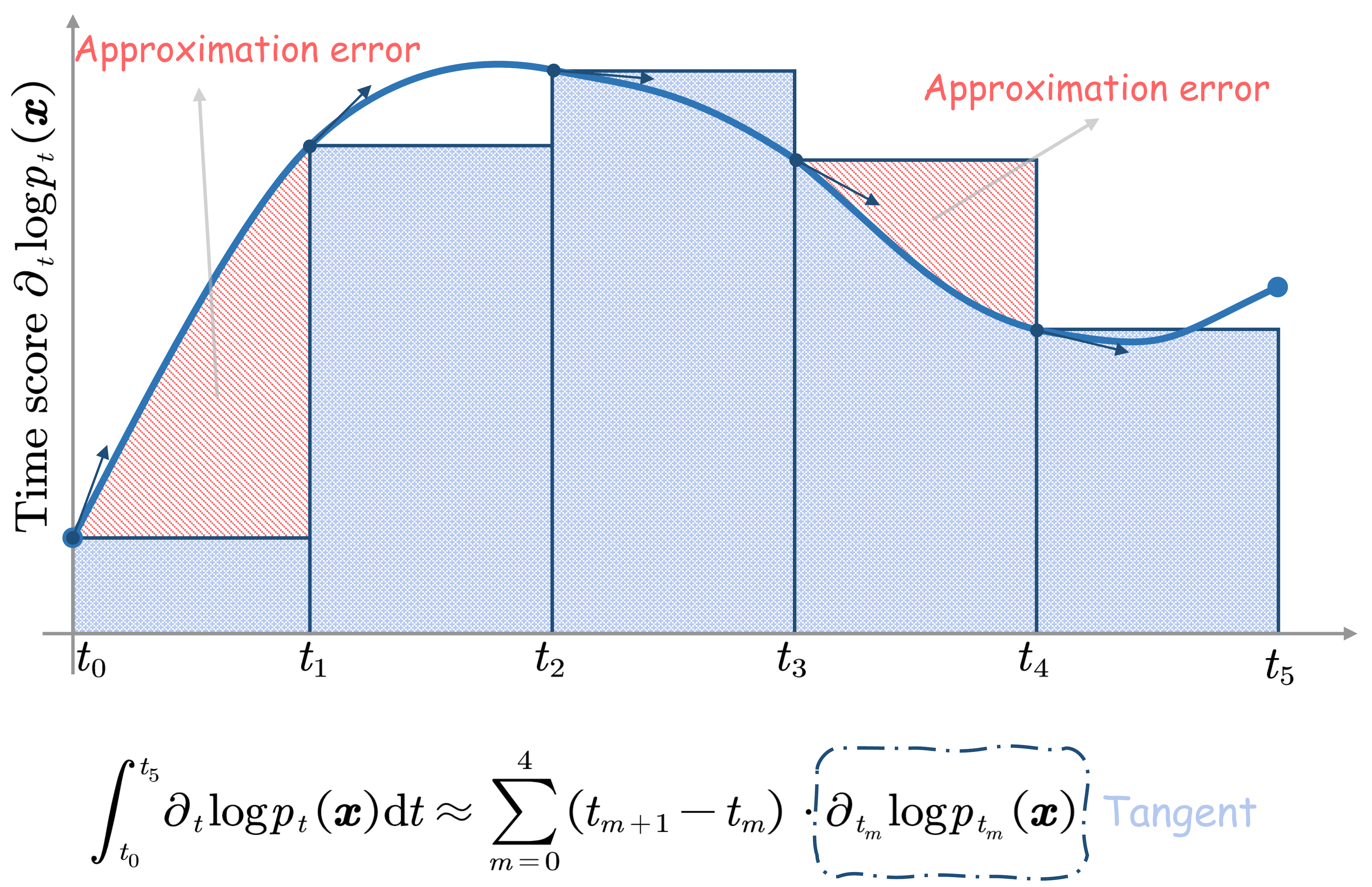}%
        }%
        \hfill
        \subfloat[Direct interval integral. \label{fig:illustration-secant}]{%
          \includegraphics[width=0.32\linewidth]{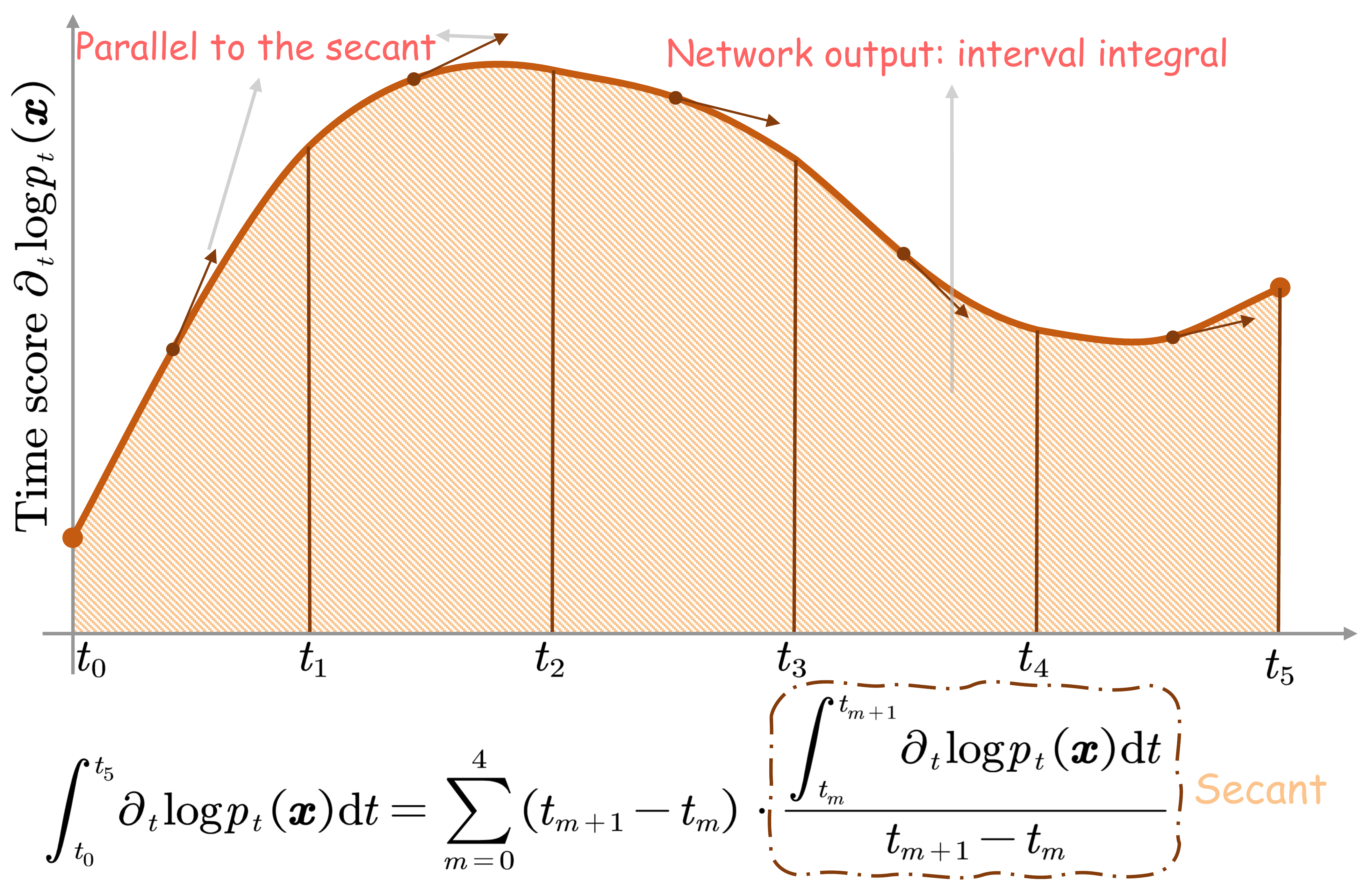}%
        }%
        \hfill
        \subfloat[Tangent = limit of secant.\label{fig:secant-alignment-illustration}]{%
          \includegraphics[width=0.32\linewidth]{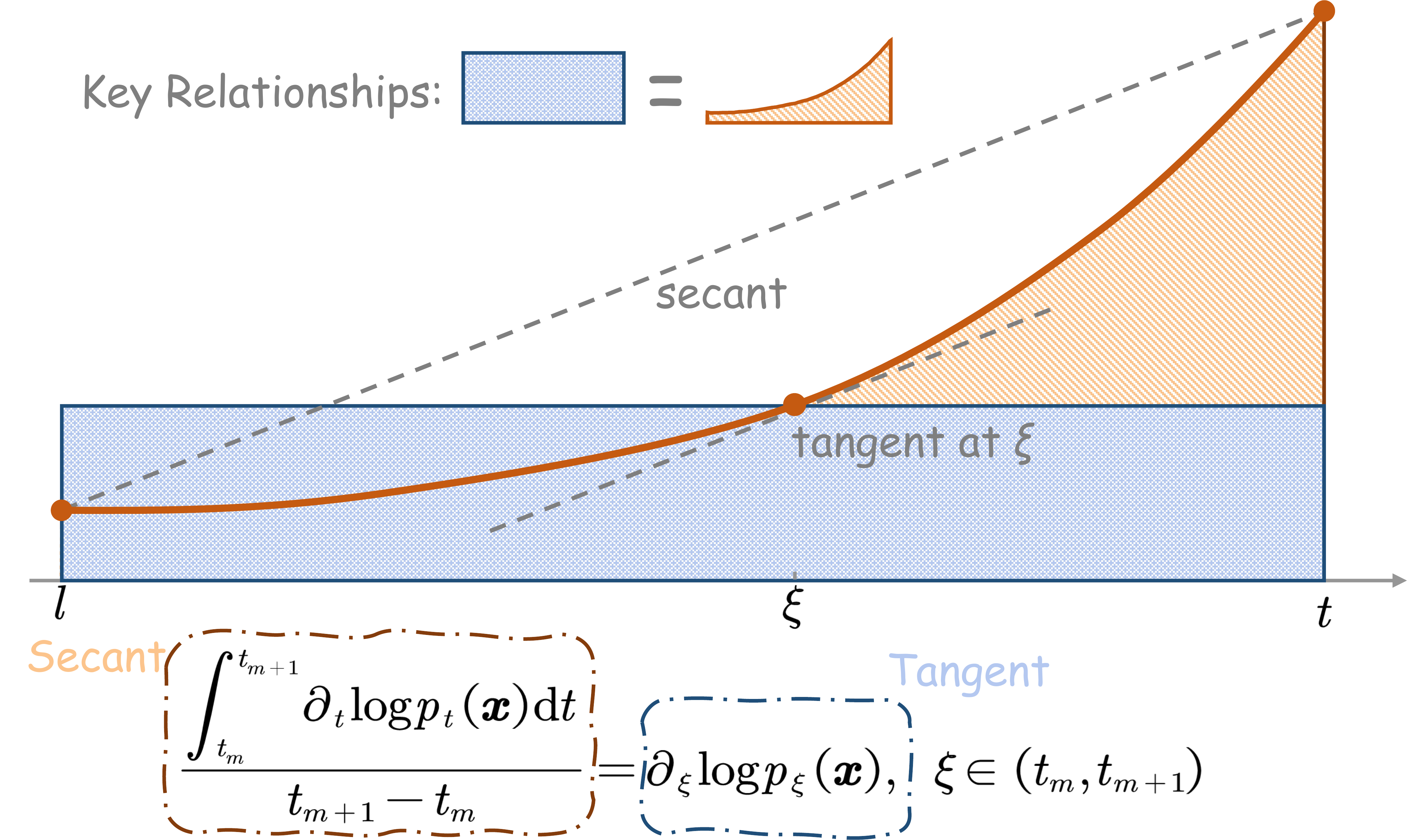}%
        }%
	
	\caption{
            Secant-based density ratio estimation generalizes tangent-based methods.
		The curve represents the time score function $\partial_t \log p_t(\boldsymbol{x})$, whose integral over $t \in [t_0, t_5]$ gives the log density ratio $\log r(\boldsymbol{x})$.
		\textbf{(a)} The tangent-based method approximates this integral by estimating the instantaneous score $\partial_t \log p_t(\boldsymbol{x})$ at discrete points and summing Riemann rectangles (blue), incurring numerical error (red hatched regions).
		\textbf{(b)} In contrast, the secant-based method directly predicts the exact integral over each sub-interval (orange shaded areas), eliminating discretization error and enabling accurate few-step inference.
		\textbf{(c)} By the mean value theorem for integral, the secant integral over $[l,t]$ equals the tangent evaluation at some $\xi \in [l,t]$; thus, the tangent method corresponds precisely to a secant method constrained to infinitesimal intervals. This establishes the tangent-based approach as a limiting case of our general secant framework. 
	}
	\label{fig:illustration-comparison-methods}
\end{figure*}

Crucially, we show that the secant offers strictly better theoretical properties: it has lower variance  (\cref{thm:variance-reduction-of-secant}), is smoother  (\cref{prop:smoothness-of-secant}), and subsumes the tangent as its infinitesimal-interval limit (see \cref{fig:illustration-comparison-methods}). This richer functional representation yields an improved optimization landscape.

We propose \textit{\textbf{I}nterval‑annealed \textbf{S}ecant \textbf{A}lignment \textbf{D}ensity \textbf{R}atio \textbf{E}stimation} (\textbf{ISA-DRE}), a unified framework applicable to \emph{both diffusive and diffusion-bridge interpolants}. Our contributions are as follows:
\begin{itemize}
    \item We propose a shift from learning instantaneous tangents to learning interval-averaged secants. We prove that the secant is a provably lower-variance (\cref{thm:variance-reduction-of-secant}), smoother (\cref{prop:smoothness-of-secant}), and strictly more general functional representation than the tangent. This yields a substantially more stable optimization landscape.

    \item We introduce the \emph{Secant Alignment Identity} (SAI) to enforce self-consistency between secants and tangents, and \emph{Contraction Interval Annealing} (CIA) to prevent bootstrap divergence by gradually expanding the learnable interval.
    
    \item We demonstrate that ISA-DRE achieves both robustness and competitive accuracy through stability. Across challenging high-discrepancy regimes, ISA-DRE achieves state-of-the-art (SOTA) accuracy with minimal function evaluations (NFE), validating stability-first learning along both diffusive and bridge-based interpolants.
    
\end{itemize}

\section{Related Works}

\subsection{Density Ratio Estimation (DRE)}
DRE research has followed two main directions: kernel-based parametric methods (Kernel DRE) and neural network–based discriminative methods (Neural DRE). Early kernel methods, such as KLIEP \cite{sugiyama2008direct} and uLSIF \cite{kanamori2009least}, minimize a Bregman or $f$-divergence \cite{sugiyama2012density} between true and estimated ratios, yielding convex optimization problems. However, they require inverting large kernel matrices in RKHS, which can become ill-conditioned in high-discrepancy or complex data, causing severe numerical instability and sensitivity to regularization \cite{kanamori2009condition}, limiting their practicality in modern deep learning.

To overcome these limitations, the field shifted to Neural DRE. Theoretical connections between DRE and binary classification (CPE) \cite{menon2016linking} showed that minimizing proper composite losses (e.g., logistic) yields consistent density ratio estimators. Methods like noise-contrastive estimation (NCE) \cite{gutmann2012noise} train discriminators to distinguish samples from $p_0$ and $p_1$, leveraging deep networks’ nonlinear power to achieve strong expressivity and improved numerical stability, avoiding the matrix inversion bottlenecks of Kernel DRE.
Despite their scalability, Neural DRE methods remain vulnerable to the density-chasm problem \cite{rhodes2020telescoping, chen2025dequantified}, where the supports of $p_0$ and $p_1$ barely overlap, leading to unstable optimization and large errors \cite{kitazawa2025bounds}. Early remedies, such as Trimmed DRE \cite{liu2017trimmed}, discard samples from high-discrepancy regions. 

Recent studies address this limitation along two directions.
The first focuses on optimization and loss refinement. In kernel-based DRE, strategies include regularized Bregman divergence for adaptive tuning \cite{zellinger2023adaptive}, iterated regularization to reduce error saturation \cite{gruber2024overcoming}, and direct ratio regularization for smoother estimates \cite{nguyen2024regularized}. Neural DRE approaches improve robustness via novel losses, e.g., $\gamma$-DRE \cite{nagumo2024density} uses $\gamma$-divergence, and new binary losses \cite{zellinger2025binary} reduce bias for large ratios. Projection Pursuit DRE \cite{wang2025projection} further mitigates the curse of dimensionality by estimating ratios in informative low-dimensional subspaces.

The second direction exploits path decomposition and continuous flows. F-DRE \cite{choi2021featurized} maps distributions into a shared feature space before estimation. Discrete decompositions, such as TRE \cite{rhodes2020telescoping}, factorize the ratio into intermediate terms, with AM-DRE \cite{xiao2022adaptive} and MDRE \cite{srivastava2023estimating} refining this idea, and IMDRE \cite{kimura2025density} generalizing interpolation along statistical-geodesic paths. Continuous flow methods, like DRE-$\infty$ \cite{choi2022density}, define a time-dependent score along an interpolation path, turning DRE into a continuous integration problem. However, the high-variance target causes unstable training and costly numerical integration at inference. Later variants, such as $\text{D}^3\text{RE}$ \cite{chen2025dequantified} and conditional-path DRE \cite{yu2025density}, improve stability but still estimate the same high-variance tangent. In contrast, we address the root cause by proposing a provably lower-variance alternative.

\subsection{Learning Integrated Functions}
A similar challenge appears in generative modeling, where diffusion models and neural ordinary differential equations (ODE) must integrate dynamics over continuous paths. 
Direct supervision on instantaneous derivatives often produces high-variance targets, causing unstable training and inefficient sampling \cite{shen2025information, li2025evodiff}. To address this, some works adopt information-theoretic formulations, either tightening likelihood bounds via KL–Fisher extensions \cite{shen2025information} or minimizing conditional variance to improve reconstruction stability \cite{li2025evodiff}.

A more effective trend is to learn integrated or interval-averaged dynamics instead of noisy instantaneous ones. MeanFlow \cite{geng2025mean} predicts average velocity rather than instantaneous velocity, while \cite{liu2025learning} introduce secant-based objectives that replace the tangent with its integral. Both demonstrate that integral-based supervision yields lower variance and more reliable learning.
Other work improves stability through structured integration and regularization. Examples include reducing ODE trajectory curvature for faster sampling \cite{lee2023minimizing}, applying model-order reduction to lower integration cost \cite{lehtimaki2022accelerating}, incorporating delays for improved robustness \cite{ji2024trainable}, and enforcing physical constraints via advection–diffusion dynamics \cite{cui2023knowledge}.

Overall, these studies support a consistent principle: integral representations provide more stable and efficient learning than high-variance instantaneous derivatives. Unlike generative models that benefit from teacher supervision or distillation \cite{liu2025learning, geng2025mean}, DRE has no ground-truth target. This motivates our ISA-DRE framework, which learns the log-density integral directly via self-supervised objectives (SAI and CIA), achieving stable optimization, low variance, and efficient inference.

\section{Background}
Let $p_0(\boldsymbol{x})$ and $p_1(\boldsymbol{x})$ be two probability density functions. The goal of DRE is to estimate the ratio $r(\boldsymbol{x}) = p_1(\boldsymbol{x})/p_0(\boldsymbol{x})$ given only samples from the two distributions. A major difficulty arises when their supports have limited overlap \cite{liu2017trimmed}.

Recent advances in score-based DRE address this challenge by introducing a continuous path that connects the two endpoint distributions. Early work such as TRE \cite{rhodes2020telescoping} adopts a discrete divide-and-conquer strategy, constructing $M$ intermediate distributions based on a linear interpolant,
\begin{equation}
	\mathbf{x}_{m/M} = \sqrt{1-\beta_{m/M}^{2}}\mathbf{x}_0 + \beta_{m/M} \mathbf{x}_1, \label{eq:interpolant-TRE}
\end{equation}
where $\mathbf{x}_0 \sim p_0$, $\mathbf{x}_1 \sim p_1$, and $\{\beta_{m/M}\}$ is an increasing sequence. The overall ratio is a telescoping product, $r(\boldsymbol{x}) = \prod_{m=0}^{M-1} r_{m/M}(\boldsymbol{x})$, but this requires training $M$ separate models and often fails to fully bridge the density gap when $M$ is small.

DRE-$\infty$ \cite{choi2022density} extends this idea by letting $M \to \infty$ and defining a  \textit{diffusion interpolant} (DI),
\begin{equation}
	\mathbf{x}_{t} = \alpha_t \mathbf{x}_0 + \beta_t \mathbf{x}_1 \quad \text{for } t \in [0,1], \label{eq:deterministic-interpolation}
\end{equation}
where coefficients $\alpha_t, \beta_t$ ensure that $\mathbf{x}_t$ smoothly transitions from being distributed as $p_0$ at $t=0$ to $p_1$ at $t=1$. The log-density ratio is then estimated via:
\begin{equation}
	\log r(\boldsymbol{x}) = \int_0^1 \partial_{t} \log p_{t}(\boldsymbol{x}) \mathrm{d}{t} = \int_0^1 s^{(t)}(\boldsymbol{x}, t) \mathrm{d}{t},
\end{equation}
where $s^{(t)}(\boldsymbol{x}, t) \triangleq \partial_{t} \log p_{t}(\boldsymbol{x})$ is the \textit{time  score} function. To enhance robustness and stability, D$^3$RE \cite{chen2025dequantified} introduces the \textit{dequantified diffusion-bridge interpolant} (DDBI), which stabilizes the path with Gaussian noise,
\begin{equation}
	\mathbf{x}_t=\alpha_t\mathbf{x}_0+\beta_t\mathbf{x}_1+\sqrt{t(1-t)\gamma^2+(\alpha_t^2+\beta_t^2)\varepsilon}\mathbf{z}, \label{eq:diffusion-bridge-interpolant}
\end{equation}
where $\mathbf{z}\sim\mathcal{N}(\mathbf{0},\boldsymbol{I}_d)$, $\gamma\in\mathbb{R}_{\geq0}$ is the noise factor, and $\varepsilon$ is a small number for stability.  DDBI behaves diffusively in the interior of the path while retaining controlled endpoints.

In all these frameworks, a neural score model  $s_{\boldsymbol{\theta}}^{(t)}$ is trained to approximate the true score $s^{(t)}$ by minimizing a time score-matching (TSM) loss \cite{choi2022density}. Since the marginal score $s^{(t)}$ is intractable, training instead uses the equivalent conditional time score-matching (CTSM) objective \cite{yu2025density},
\begin{equation} \label{eq:conditional-time-score-matching}
	\mathcal{L}(\boldsymbol{\theta})=\mathbb{E}
	\left[\lambda(t)\left|s^{(t)}(\boldsymbol{x}_t,t\mid \boldsymbol{y}) - s_{\boldsymbol{\theta}}^{(t)}(\boldsymbol{x}_t,t)\right|^{2}\right],       
\end{equation}
where $\lambda(t)= 1/\mathrm{Var}_{p_t}(s^{(t)}\mid \boldsymbol{y})$ is a weighting function, the conditioning variable $\boldsymbol{y}$ is $\boldsymbol{x}_1$ for DI and $(\boldsymbol{x}_0, \boldsymbol{x}_1)$ for DDBI, and the conditional score $s^{(t)}(\boldsymbol{x}_t,t\mid \boldsymbol{y})=\partial_t\log p_{t}(\boldsymbol{x}_t\mid \boldsymbol{y})=$ 
\begin{equation}\label{eq:conditional-time-score}
	\begin{cases}
		- \frac{d\dot{\alpha}_t}{\alpha_t} + \frac{\dot{\alpha}_t \|\boldsymbol{x}_t-\beta_t \boldsymbol{x}_1\|^2}{\alpha_t^3}  + \frac{\dot{\beta}_t (\boldsymbol{x}_t-\beta_t \boldsymbol{x}_1)^\top \boldsymbol{x}_1}{\alpha_t^2}, & \text{if DI},\\
		-\frac{d\dot{\sigma}_t^2}{2\sigma_t^2} + \frac{\|\boldsymbol{x}_t - \boldsymbol{\mu}_t\|^2}{2(\sigma_t^2)^2}(\dot{\sigma}_t^2) + \frac{(\boldsymbol{x}_t - \boldsymbol{\mu}_t)^\top \dot{\boldsymbol{\mu}}_t}{\sigma_t^2}, & \text{if DDBI},
	\end{cases}
\end{equation}
where $\boldsymbol{\mu}_t = \alpha_t\boldsymbol{x}_0 + \beta_t\boldsymbol{x}_1$, $\sigma_t^2 = t(1-t)\gamma^2 + (\alpha_t^2 + \beta_t^2)\varepsilon$, and dots denote time derivatives (e.g., $\dot{\alpha}_t = \frac{\mathrm{d} \alpha_t}{\mathrm{d} t}$).

After training, the ratio is estimated by numerically integrating the learned score: $\log r_{\boldsymbol{\theta}^\star}(\boldsymbol{x}) = \int_0^1 s_{\boldsymbol{\theta}^\star}^{(t)} (\boldsymbol{x}, t)\mathrm{d}t$.

\section{Methods}
This section details the underlying theory and practical implementation of our ISA-DRE framework.

\subsection{A Low-Variance and Smooth Learning Target}
\label{sec:secant-alignment}

\subsubsection{The Secant Function}
ISA-DRE is based on the \textbf{\textit{secant}} function $u$, which is the average of the time score function $s^{(t)}$ (a.k.a. the tangent function) over a time interval $[l, t]$,
\begin{equation}
	u(\boldsymbol{x}_t, l, t) \triangleq 
	\begin{cases}
		s^{(t)}(\boldsymbol{x}_t, t), & \text{if } l = t, \\
		\displaystyle \frac{1}{t - l} \int_{l}^{t} s^{(t)}(\boldsymbol{x}_{\tau}, \tau) \mathrm{d}\tau, & \text{if } l \neq t.
	\end{cases}
\end{equation}
In the limiting case $l = t$, we define $u(\boldsymbol{x}_t, t, t) = s^{(t)}(\boldsymbol{x}_t, t)$ since $u$ is continuous at $l=t$, i.e., 
\begin{equation}
	u(\boldsymbol{x}_t, t, t) =\lim_{l\to t} \frac{1}{l - t} \int_{l}^{t} s^{(t)}(\boldsymbol{x}_\tau, \tau) \mathrm{d}\tau=s^{(t)}(\boldsymbol{x}_t, t).
\end{equation}

This definition reveals the relationship between the secant and the tangent functions:
\begin{equation}\label{eq:tangent-secant-relationship}
	u(\boldsymbol{x}_t, l, t) = \mathbb{E}_{p(\tau)}\left[ s^{(t)}(\boldsymbol{x}_\tau, \tau) \right]= s^{(t)}(\boldsymbol{x}_\xi, \xi), 
\end{equation}
where $p(\tau)=\mathcal{U}[l, t]$ and $\xi\in[l, t]$. The first equality follows by definition, and the second by the mean value theorem for integrals, assuming $s^{(t)}$ is continuous on $[l, t]\subseteq[0,1]$.

The equations in \cref{eq:tangent-secant-relationship} yield two insights: (1) The secant over the interval $[l, t]$ equals the expectation of all tangents on that interval; and (2) it can also be viewed as a reparameterized tangent evaluated at some  $\xi\in[l, t]$.
A geometric illustration of (2) is provided in \cref{fig:secant-alignment-illustration}. Leveraging (1), we argue that the secant offers more favorable learning properties.

\subsubsection{Theoretical Justification on Stability and Smoothness}
Our motivation originates from the instability of tangent-based methods: learning the tangent $s^{(t)}$ is inherently high-variance. In contrast, learning its integral, the secant $u$, is a more stable objective. This intuition is formalized in the following theorem.

\begin{restatable}[Low-Variance Secant Target]{theorem}{VarianceSecantTangent} 
	\label{thm:variance-reduction-of-secant}
	Let $l$ and $t$ be independent random variables with joint probability density $p(l,t)$ on $[0,1]^2$, conditioned on $l \leq t$.
	For a fixed data point $\boldsymbol{x} \sim p_1$, define the secant variable $U \triangleq u(\boldsymbol{x}, l, t)$ and tangent variable $S \triangleq s^{(t)}(\boldsymbol{x}, \tau)$, where $\tau \sim p(\tau)=\mathcal{U}[l, t]$. Under the joint distribution $p(l, t)$, the variance of $U$ w.r.t. $p(l, t)$ satisfies:
	\begin{equation}
		\mathrm{Var}_{p(l, t)}(U) \leq \mathrm{Var}_{p(\tau)}(S),
	\end{equation}
	with equality $\mathrm{iff}$ $S$ is constant for $p$-almost every $\tau \in [0,1]$. 
\end{restatable}

\begin{remark}
	We note that \cref{thm:variance-reduction-of-secant} is stated for a fixed $\boldsymbol{x}$ to isolate the variance reduction from temporal averaging. This principle extends to the full training distribution where $\boldsymbol{x}_\tau$ varies with $\tau\sim \mathcal{U}[l,t]$. The low-variance property of the secant $u(\boldsymbol{x}, l, t \mid \boldsymbol{y}) = \frac{1}{t-l} \int_{l}^{t} s^{(t)}(\boldsymbol{x}_{\tau}, \tau \mid \boldsymbol{y}) \mathrm{d}\tau$ holds for any given $\boldsymbol{y}$. By the law of total variance, the total variance of the secant target (averaged over both time $\tau$ and data $\boldsymbol{y}$) remains provably lower than the total variance of the tangent target.
\end{remark}

See \cref{proof:variance-reduction-of-secant} for a proof. \cref{thm:variance-reduction-of-secant} is the cornerstone of our approach. It proves that the secant $u$ is a provably lower-variance target  than the tangent $s^{(t)}$ itself. This suggests that a model $u_{\boldsymbol{\theta}}$ trained to approximate the secant will experience a more stable optimization landscape, particularly in high-discrepancy density-chasm scenarios where individual tangent estimates can be extremely noisy.

Furthermore, this averaging process yields a target function that is better behaved and smoother than its tangent-based counterpart, as shown in the following proposition.
\begin{restatable}[Smoothness of the Secant Function]{proposition}{SmoothnessSecantFunction}
	\label{prop:smoothness-of-secant}
	Assume the score function $s^{(t)}(\boldsymbol{x}, \tau)$ is $\lambda$-Lipschitz continuous in $\tau$.
	Then for fixed $\boldsymbol{x}$ and $l \in [0,1)$, the secant function $u(\boldsymbol{x}, l, t)$ is $\frac{\lambda}{2}$-Lipschitz continuous in $t$.
\end{restatable}
See \cref{proof:smoothness-of-secant} for a proof. \cref{thm:variance-reduction-of-secant} and \cref{prop:smoothness-of-secant} together establish that the secant $u$ is a statistically more stable and mathematically smoother learning target than the tangent $s^{(t)}$. The central goal of our method is therefore to learn this superior target function $u$ directly.

\subsubsection{The Secant Alignment Identity (SAI)}
To avoid explicit time integration, we derive a differential identity linking the secant function to the readily evaluable conditional tangent function.
Adopting a Lagrangian perspective, we track the time score along trajectories of a time-dependent interpolant $\boldsymbol{x}_\tau \sim p_\tau$. This generalizes the inference setting, where $\boldsymbol{x}_\tau \equiv \boldsymbol{x}$ for all $\tau \in [0,1]$, to allow  $\boldsymbol{x}_\tau$ to evolve with $\tau$. 

By rearranging the integral definition of $u$ as,
\begin{equation}
    (t - l) u(\boldsymbol{x}_t, l, t) = \int_{l}^{t} s^{(t)}(\boldsymbol{x}_\tau, \tau) \mathrm{d}\tau, 
\end{equation}
training proceeds on samples $\boldsymbol{x}_t$ moving along this path. Differentiating both sides w.r.t. $t$ using the \textit{total derivative} $\frac{\mathrm{d}}{\mathrm{d}t}$, which accounts for both explicit time dependence and motion via  $\frac{\mathrm{d}\boldsymbol{x}_t}{\mathrm{d}t}$ yields the \textbf{\textit{Secant Alignment Identity}} (\textbf{SAI}):
\begin{equation}
	\underbrace{u(\boldsymbol{x}_t, l, t)}_{\text{Secant Function}} = \underbrace{s^{(t)}(\boldsymbol{x}_t, t)}_{\text{Tangent Function}} - \underbrace{(t - l)\cdot \frac{\mathrm{d}}{\mathrm{d}t}u(\boldsymbol{x}_t, l, t)}_{\text{Correction Term}},
\end{equation}
where $\frac{\mathrm{d}}{\mathrm{d}t} u$ is the \textit{total derivative} of $u$ w.r.t. $t$. 
Since $\frac{\mathrm{d}l}{\mathrm{d}t}=0$ and $\frac{\mathrm{d}t}{\mathrm{d}t}=1$, this derivative can be derived via the chain rule:
\begin{equation}
	\begin{aligned}
		\frac{\mathrm{d}}{\mathrm{d}b} u(\boldsymbol{x}_t, l, t) &= \frac{\mathrm{d} \boldsymbol{x}_t}{\mathrm{d}t} \cdot \partial_{\boldsymbol{x}} u + \frac{\mathrm{d}l}{\mathrm{d}t} \cdot \partial_l u + \frac{\mathrm{d}t}{\mathrm{d}t} \cdot \partial_t u  \\
		&=\frac{\mathrm{d} \boldsymbol{x}_t}{\mathrm{d}t} \cdot \partial_{\boldsymbol{x}} u + \partial_t u.
	\end{aligned}
\end{equation}

Here, the derivative term $\frac{\mathrm{d} \boldsymbol{x}_t}{\mathrm{d}t}$ is known analytically from the chosen interpolant, e.g., \cref{eq:deterministic-interpolation} or \cref{eq:diffusion-bridge-interpolant},
\begin{equation}
	\frac{\mathrm{d} \boldsymbol{x}_t}{\mathrm{d}t}=\begin{cases}
		\frac{\mathrm{d} \alpha_t}{\mathrm{d}t}\boldsymbol{x}_0 + \frac{\mathrm{d} \beta_t}{\mathrm{d}t}\boldsymbol{x}_1, & \text{ if DI},\\
		\frac{\mathrm{d} \alpha_t}{\mathrm{d}t}\boldsymbol{x}_0 + \frac{\mathrm{d} \beta_t}{\mathrm{d}t}\boldsymbol{x}_1+\frac{\gamma(1-2t)}{2\sqrt{t(1-t)}}\boldsymbol{z}, & \text{ if DDBI},
	\end{cases}
\end{equation}
where $\boldsymbol{z}\sim \mathcal{N}(\mathbf{0}, \boldsymbol{I}_d)$, and $\frac{\mathrm{d} \alpha_t}{\mathrm{d}t}$, $\frac{\mathrm{d} \beta_t}{\mathrm{d}t}$ denote time derivatives of the coefficients.
The partial derivatives $\partial_{\boldsymbol{x}} u$ and $\partial_t u$ are evaluated using the Jacobian-vector product (JVP) between the Jacobian $[\partial_{\boldsymbol{x}} u, \partial_l u, \partial_t u]$ and the direction vector $\left[\frac{\mathrm{d} \boldsymbol{x}_t}{\mathrm{d}t}, 0, 1 \right]$.

The SAI thus provides a self-consistency condition: the secant over $[l, t]$ is recoverable from the tangent at $t$ and a correction term, enabling direct training of $u$.

\subsubsection{Training with the SAI}
The theoretical benefits of the secant target (\cref{thm:variance-reduction-of-secant}) and the tractability offered by the SAI provide a clear prescription for training a neural network $u_{\boldsymbol{\theta}}$ to approximate the true secant function $u$.

We replace the parameterized tangent term in the CTSM objective (\cref{eq:conditional-time-score-matching}) with its SAI-based representation, yielding the Conditional Secant Alignment (CSA) loss:
\begin{equation} \label{eq:conditional-secant-alignment-loss}
	\mathcal{L}_{\text{CSA}}(\boldsymbol{\theta})=\mathbb{E}\left[\lambda(t)\left|s^{(t)}(\boldsymbol{x}_t, t\mid \boldsymbol{y}) - s_{\boldsymbol{\theta}}^{(t)}(\boldsymbol{x}_t, t)\right|^{2}\right],     
\end{equation}
where $s_{\boldsymbol{\theta}}^{(t)}(\boldsymbol{x}_t,t) = u_{\boldsymbol{\theta}}(\boldsymbol{x}_t, l, t) + \text{sg}\left( (t - l) \frac{\mathrm{d}}{\mathrm{d}t}u_{\boldsymbol{\theta}}(\boldsymbol{x}_t, l, t) \right)$ and $\text{sg}$ denotes a stop-gradient operation to prevent feedback amplification, consistent with consistency training practice \cite{song2024improved}.

The expectation is over $(l,t)$, $(\boldsymbol{x}_0,\boldsymbol{x}_1)$, the interpolant $\boldsymbol{x}_t$, and the conditioning variable $\boldsymbol{y}$. Minimizing $\mathcal{L}_{\text{CSA}}$ trains  $u_{\boldsymbol{\theta}}$ to be consistent with the true secants across all sub-intervals. Training and inference details appear in \cref{alg:secant_alignment}.

Finally, \cref{proposition:secant-tangent-consistency} provides theoretical support: the SAI condition defines a first-order linear relation in $t$, and under mild regularity assumptions, the desired secant function is the unique solution given boundary conditions.

\begin{restatable}{proposition}{SecantTangentConsistency}
	 \label{proposition:secant-tangent-consistency}
	Let the time score function $s^{(t)}(\boldsymbol{x}, \tau)$ be continuous in $\tau$, and the secant model $u_{\boldsymbol{\theta}}(\boldsymbol{x}, l, t)$ be continuously differentiable in $t$.
	Then, the learned secant $u_{\boldsymbol{\theta}^{\star}}$ exactly matches the true secant:
	\begin{equation}
		u_{\boldsymbol{\theta}^{\star}}(\boldsymbol{x}, l, t) = u(\boldsymbol{x}, l, t)= \frac{1}{t-l} \int_l^t s^{(t)}(\boldsymbol{x}, \tau) \mathrm{d}\tau,
	\end{equation}
	for all $(\boldsymbol{x}, l, t)$ with $l \neq t$, if and only if the following hold: $(1)$  Boundary condition: $\lim_{t \to t_0} u_{\boldsymbol{\theta}^{\star}}(\boldsymbol{x}, t_0, t) = s^{(t)}(\boldsymbol{x}, t_0)$ for any fixed $t_0 \in [0,1]$. $(2)$ Consistency condition (i.e., SAI): $s^{(t)}(\boldsymbol{x}, t) = u_{\boldsymbol{\theta}^{\star}}(\boldsymbol{x}, l, t) + (t-l)\frac{\mathrm{d}}{\mathrm{d}t} u_{\boldsymbol{\theta}^{\star}}(\boldsymbol{x}, l, t)$.
\end{restatable}
Proof can be found in Appendix \ref{proof:secant-tangent-consistency}. 
This result establishes that enforcing the SAI is a necessary and sufficient condition for consistency, identifying the true secant as the unique fixed point of our learning objective. We note, however, that this proposition characterizes the properties of the optimal solution rather than the convergence dynamics of the optimization algorithm itself. Under the assumption that the training procedure successfully reaches this fixed point, conditions (1) and (2) ensure that $u_{\boldsymbol{\theta}^{\star}}$ produces the correct $u$ at inference time, thereby validating the secant alignment framework.

\subsubsection{Inference via Flexible-Step Estimation}
By the fundamental theorem of calculus, the secant function $u$ satisfies $\log p_{t}(\boldsymbol{x}) - \log p_{l}(\boldsymbol{x})
= (t - l)u(\boldsymbol{x},l,t)$.
Setting $l=0$ and $t=1$ yields a direct log-density ratio estimate for  $\boldsymbol{x}\sim p_1$,
\begin{equation} \label{eq:one-step-estimation}
	\log r(\boldsymbol{x})
	= \log p_1(\boldsymbol{x}) - \log p_0(\boldsymbol{x})
	= u(\boldsymbol{x},0,1),
\end{equation}
so that once $u_{\boldsymbol{\theta}^\star}$ is trained, inference becomes
\begin{equation}
	\log r_{\boldsymbol{\theta}^\star}(\boldsymbol{x})
	= u_{\boldsymbol{\theta}^\star}(\boldsymbol{x},0,1), \boldsymbol{x}\sim p_1.
\end{equation}

Although the $M=1$ estimator in \cref{eq:one-step-estimation} is theoretically sound, in practice we find that a multi-step estimate ($M>1$) provides a  flexible balance between accuracy and efficiency:
\begin{equation} \label{eq:multi-step-estimation}
	\begin{aligned}
		\log r_{\boldsymbol{\theta}^\star}(\boldsymbol{x})
		&= \sum_{m=0}^{M-1} \left(\log p_{t_{m+1}}(\boldsymbol{x}) - \log p_{t_m}(\boldsymbol{x})\right) \\
		&= \sum_{m=0}^{M-1} (t_{m+1}-t_m)u_{\boldsymbol{\theta}^\star}(\boldsymbol{x},t_m,t_{m+1}),
	\end{aligned}
\end{equation}
where the sequence $0=t_0<t_1<\cdots<t_M=1$ is a partition of the interval $[0,1]$.
$M$ corresponds to the \textit{number of function evaluations} (\textbf{NFE}) as defined in \cite{chen2018neural}.

The key advantage of ISA-DRE is not merely enabling $M=1$ inference, but the quality of the learned function $u_{\boldsymbol{\theta}^\star}$. Because it is trained to approximate a smooth and low-variance integral target (per \cref{thm:variance-reduction-of-secant} and \cref{prop:smoothness-of-secant}), $u_{\boldsymbol{\theta}^\star}$ provides a stable and accurate estimate of the log-density integral—unlike tangent-based models, which must numerically integrate noisy instantaneous predictions $s_{\boldsymbol{\theta}^\star}$.

As a result, ISA-DRE achieves high accuracy with very small NFE (e.g., $M=5$), summing high-fidelity interval estimates rather than averaging out high-variance derivative noise. This enables efficient and real-time density ratio estimation without reliance on costly ODE solvers or dense quadrature schemes.

\subsection{Stable Training via Contraction Interval Annealing}
\label{sec:cia}

\subsubsection{Bootstrap Divergence in SAI}
Although the secant target $u$ is statistically stable (\cref{thm:variance-reduction-of-secant}), the optimization induced by the CSA loss (\cref{eq:conditional-secant-alignment-loss}) introduces a separate challenge, which we term \emph{bootstrap divergence}.

\begin{figure}[ht]
	\centering
	\includegraphics[width=0.8\linewidth]{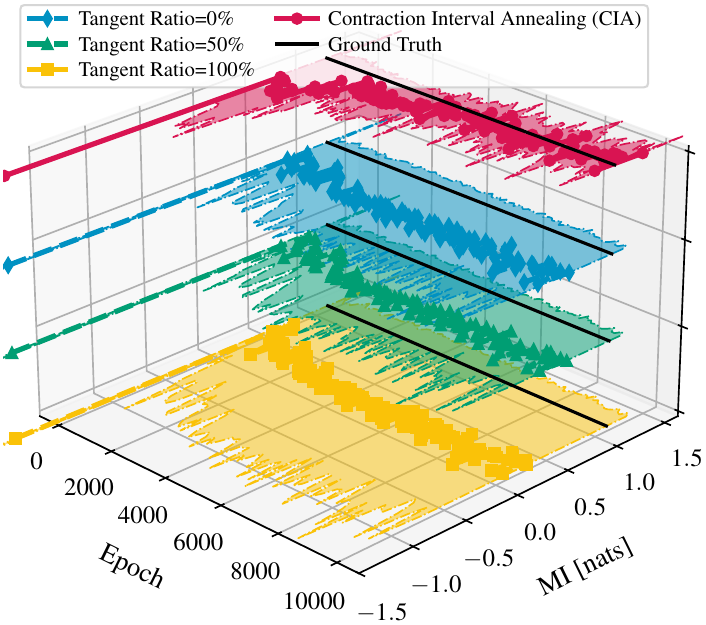}  
	\caption{Mutual information estimation on the $\mathsf{AdditiveNoise}$ dataset with CIA and fixed tangent ratios. The tangent ratio denotes the proportion of samples with $l = t$, corresponding to tangent-only ($100\%$) or secant-only ($0\%$) supervision (see \cref{section:practical-choices}). Shaded areas show ``std'' across samples. CIA ensures \textit{stable and consistent} convergence.
	}
	\label{fig:ablation-IA-epoch}
\end{figure}

In CSA, the secant model $u_{\boldsymbol{\theta}}$ is supervised by a target $s_{\boldsymbol{\theta}}^{(t)}$ that depends on the model's own temporal derivative $\tfrac{\mathrm{d}}{\mathrm{d}t}u_{\boldsymbol{\theta}}$. When the interval length $|t-l|$ is large, the correction term $(t-l)\tfrac{\mathrm{d}}{\mathrm{d}t}u_{\boldsymbol{\theta}}$ tends to amplify estimation errors early in training, leading to feedback loops and divergence. This instability is empirically evident in \cref{fig:ablation-IA-epoch}, where training with secant-only supervision (0\% tangent ratio) fails to converge.

\subsubsection{Heuristic Motivation for Interval Annealing}
We provide a stability analysis to motivate our annealing schedule based on contraction mapping principles. Formally, define the operator $\mathcal{T}$ implicitly associated with the SAI fixed-point iteration as:
\begin{equation}
	\mathcal{T}(u) \triangleq s^{(t)} - (t - l) \frac{\mathrm{d}}{\mathrm{d}t}u.
\end{equation}

Training seeks a function $u$ such that $u \approx \mathcal{T}(u)$. From the perspective of the Banach fixed-point theorem, the stability of this iteration is governed by the contraction factor $C$:
\begin{equation}
	\|\mathcal{T}(u_1) - \mathcal{T}(u_2)\| \le C \|u_1 - u_2\|.
\end{equation}

In a suitable normed space, this factor $C$ is proportional to the interval length $|t-l|$ and the norm of the derivative operator $\tfrac{\mathrm{d}}{\mathrm{d}t}$: $C \propto |t - l| \cdot \left\| \tfrac{\mathrm{d}}{\mathrm{d}t} \right\|$.

A key insight concerns the function space in which neural networks operate. While the differentiation operator is unbounded in $L^2([0,1])$, neural networks effectively restrict the hypothesis class to smoother functions due to architectural biases and regularization. In such effectively smooth spaces (e.g., bounded Sobolev norms), the derivative operator has a finite norm $K \approx \left\| \tfrac{\mathrm{d}}{\mathrm{d}t} \right\|$. The stability condition thus suggests requiring $|t - l| \cdot K < 1$. This relationship explicates the source of instability: large intervals ($|t-l| \to 1$) coupled with initial training dynamics (where $K$ may fluctuate) can cause the contraction constant to exceed $1$, violating the stability condition and triggering bootstrap divergence.

\subsubsection{Contraction Interval Annealing (CIA)}
To mitigate this issue, we introduce \emph{Contraction Interval Annealing (CIA)}, a curriculum strategy that dynamically adjusts the secant interval length $|t-l|$ from $0$ to $1$ during training.

At the onset of training, CIA sets $|t-l|\approx 0$. Consequently, the influence of the derivative correction term vanishes:
\begin{equation}
	\left\| (t - l) \frac{\mathrm{d}}{\mathrm{d}t}u_{\boldsymbol{\theta}} \right\| \leq |t-l| \left\| \frac{\mathrm{d}}{\mathrm{d}t}u_{\boldsymbol{\theta}} \right\| \approx 0.
\end{equation}
In this limit, the contraction factor approaches zero, and the CSA objective (\cref{eq:conditional-secant-alignment-loss}) effectively reduces to a well-conditioned local tangent alignment task. This provides a stable warm-up phase where the model aligns with the local vector field without requiring accurate derivative estimates.

As optimization progresses and the function $u_{\boldsymbol{\theta}}$ becomes smoother (stabilizing $K\approx \left\| \tfrac{\mathrm{d}}{\mathrm{d}t} \right\|$), CIA gradually increases $|t-l|$ towards $1$, fully restoring the long-range secant constraint. This curriculum bridges the gap between algorithmic stability and statistical efficiency, enabling the model to learn the low-variance secant target over the full interval $[0,1]$ while avoiding early-stage divergence.

\subsection{Practical Choices for Implementation}
\label{section:practical-choices}
We summarize the practical setup of ISA-DRE. Additional implementation details and ablations are provided in \cref{section:ablation-study}.

\subsubsection{Time Sampler for Interval Sampling}
We evaluate three strategies for sampling secant intervals $(l,t)$: (1) Uniform (\textbf{Uni.}), sampling with $l, t\sim\mathcal{U}(0,1)$; (2) Logit‑Normal (\textbf{LN}) sampling via a logistic transform of Gaussian noise \cite{geng2025mean}; (3) Variance‑based Importance (\textbf{VI}) sampling with  $t\sim p(t)\propto1/\mathrm{Var}_{p_t}(s^{(t)}\mid\boldsymbol{y})$, motivated by the weighting function in CTSM (\cref{eq:conditional-time-score-matching}). Each sampled pair is sorted to ensure $l \le t$.

The VI sampler naturally aligns with our stability-first goal: it allocates more probability to time regions with smaller conditional score variance, leading to faster convergence and improved robustness compared with uniform sampling.
The variance of the conditional time score admits a closed-form expression and can be computed analytically as~\cite{yu2025density}
\begin{equation}
	\mathrm{Var}_{p_t}(s^{(t)}\mid \boldsymbol{y})=\begin{cases}
		\frac{2d \dot{\alpha}_t^2}{\alpha_t^2} + \frac{\dot{\beta}_t^2}{\alpha_t^2} \|\boldsymbol{x}_1\|^2, & \text{if DI},\\
		\frac{d(\dot{\sigma}_t^2)^2}{2(\sigma_t^2)^2} + \frac{\|\dot{\alpha}_t\boldsymbol{x}_0 + \dot{\beta}_t\boldsymbol{x}_1\|^2}{\sigma_t^2}, & \text{if DDBI},
	\end{cases}
\end{equation}
where all symbols are defined as in \cref{eq:conditional-time-score}.

\subsubsection{Secant-Tangent Supervision (STS)}
As a baseline, we consider a fixed secant–tangent supervision ratio, following the spirit of \cite{geng2025mean}: a predefined fraction of samples use the tangent case ($l=t$), while the rest use secant intervals ($l\ne t$). 
In contrast, CIA gradually increases the admissible interval length $|t-l|\to 1$ over training. This adaptive schedule enables a smooth transition from stable local tangent alignment to full secant learning, avoiding the instability that arises when large intervals are applied too early.
\begin{algorithm}[h]
	\caption{Training and inference procedures for ISA-DRE}
	\label{alg:secant_alignment}
	\textbf{Input}: Secant model $u_{\boldsymbol{\theta}}$, data distributions $p_0, p_1$, time distribution $p(l, t)$, interpolant schedules, estimation step $M$.\\
	\textbf{Output}: Trained model $u_{\boldsymbol{\theta}^{\star}}$ and estimated log-ratio $\log r_{\boldsymbol{\theta}^\star}(\boldsymbol{x})$. 
    
	\begin{algorithmic}[1]
		\STATE // Training procedure for ISA-DRE
		\REPEAT
        \STATE Sample a sample pair $(\boldsymbol{x}_0, \boldsymbol{x}_1) \sim p_0\times p_1$.
		\STATE Sample a secant interval $[l,t]$ via $p(l, t)$ and CIA.  
		\STATE Get $\boldsymbol{x}_t$ via interpolant schedules using $(\boldsymbol{x}_0, \boldsymbol{x}_1)$.
		\STATE Set conditional variable $\boldsymbol{y}$ via interpolant schedules.
		\STATE Get target conditional tangent $s^{(t)} \leftarrow s^{(t)}(\boldsymbol{x}_t, t \mid \boldsymbol{y})$.   
		\STATE Estimate secant $u_{\boldsymbol{\theta}} \leftarrow u_{\boldsymbol{\theta}}(\boldsymbol{x}_t, l, t)$.   
		\STATE Estimate  $\frac{\mathrm{d}}{\mathrm{d}t} u_{\boldsymbol{\theta}} \leftarrow \text{JVP}(u_{\boldsymbol{\theta}}, (\boldsymbol{x}_t, l, t), (\frac{\mathrm{d}\boldsymbol{x}_t}{\mathrm{d}t}, 0, 1))$. 
		\STATE Estimate tangent $s^{(t)}_{\boldsymbol{\theta}} \leftarrow u_{\boldsymbol{\theta}}  + \text{sg}\left((t - l) \frac{\mathrm{d}}{\mathrm{d}t} u_{\boldsymbol{\theta}}\right)$. 
		\STATE Compute loss $\mathcal{L}_{\text{CSA}} \! \leftarrow \! \left|s^{(t)} - s^{(t)}_{\boldsymbol{\theta}}\right|^2 \! / \mathrm{Var}_{p_t}(s^{(t)}\mid \boldsymbol{y})$. 
		\STATE Update parameters $\boldsymbol{\theta}$ using $\nabla_{\boldsymbol{\theta}} \mathcal{L}_{\text{CSA}}$. 
		\UNTIL{convergence}
		\STATE // Inference procedure for ISA-DRE
		\STATE Sample a data $\boldsymbol{x}\sim p_1$.
		\STATE Estimate log density ratio $\log r_{\boldsymbol{\theta}^\star}(\boldsymbol{x})$ via \cref{eq:multi-step-estimation}.
	\end{algorithmic}
\end{algorithm}


\section{Experiments}
\subsection{Baseline Settings}

For density estimation tasks, we consider Neural DRE  methods spanning multiple objective families: (1) Classification-based losses such as NCE \cite{gutmann2010noise} and InfoNCE \cite{oord2018representation}; (2) $f$-divergence–based objectives including Pearson $\chi^2$ (neural uLSIF) \cite{kanamori2009least} and Hellinger \cite{birrell2021variational}; (3) Robust formulations such as $\gamma$-DRE \cite{nagumo2024density}, EW-DRE, and PW-DRE \cite{zellinger2025binary}.
While these discriminative objectives work well in lower dimensions, we found them inadequate for high-dimensional density-chasm scenarios. Thus, for our primary high-discrepancy benchmarks (MI estimation), we focus on integral-based estimators: tangent-based DRE-$\infty$ \cite{choi2022density}, D$^3$RE \cite{chen2025dequantified}, and our secant-based ISA-DRE. Unless otherwise specified, integrals for DRE-$\infty$ and D$^3$RE are approximated using the trapezoidal rule. ISA-DRE uses the VI sampler and employs CIA for stable training by default.

\subsection{Illustration of the Secant vs. the Tangent}
\begin{figure}[ht]
	\centering
	\includegraphics[width=0.98\linewidth]{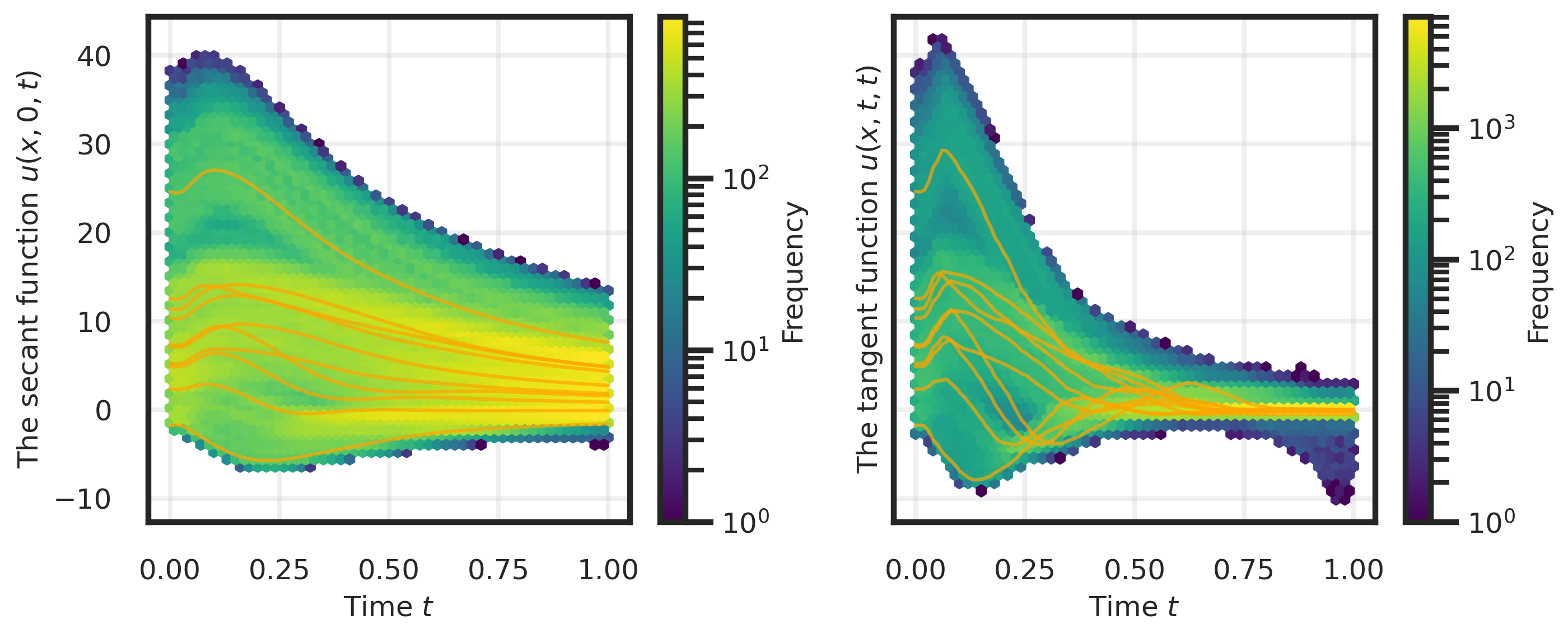}
	\caption{Comparison of the learned secant function $u_{\boldsymbol{\theta}}(\boldsymbol{x}, 0, t)$ (left) and tangent function $u_{\boldsymbol{\theta}}(\boldsymbol{x}, t, t)$ (right). Each orange curve shows $u$ over time $t$ for a fixed $\boldsymbol{x}$. The secant curves are \textit{smoother and more concentrated.}}
	\label{fig:secant-vs-tangent}
\end{figure}

To compare the learned secant and tangent functions, we visualize their trajectories in \cref{fig:secant-vs-tangent}. The left panel shows the secant  $u_{\boldsymbol{\theta}}(\boldsymbol{x}, 0, t)$ and the right shows the tangent $u_{\boldsymbol{\theta}}(\boldsymbol{x}, t, t)$, with each orange curve representing to a fixed $\boldsymbol{x}$.

The secant trajectories are noticeably smoother and more tightly concentrated, especially at early timesteps, whereas tangent trajectories fluctuate more and exhibit substantially higher variance. Secants also cluster around their mean, while tangents remain widely dispersed. This empirically confirms the stability advantage predicted by \cref{thm:variance-reduction-of-secant}.

\subsection{Results for Density Estimation}
\label{sec:density_estimation}

To assess our model’s ability to capture complex target distributions $p_{\mathrm{data}}$, we express them via a simple base distribution $p_0$ (e.g., $\mathcal{N}(\mathbf{0},\boldsymbol{I}_{d})$) using the learned density ratio. The log-likelihood is approximated by  $\log p_{\mathrm{data}}(\boldsymbol{x}) \approx \log r_{\boldsymbol{\theta}^\star}(\boldsymbol{x}) + \log p_{0}(\boldsymbol{x})$.
For all methods, we use identical network architectures to ensure fair comparison.

\subsubsection{Experimental Setup}
For all methods, we sample batches of  $(\boldsymbol{x}_0, \boldsymbol{x}_1) \sim p_0\times p_1$ and $t \sim p(t)$ over $[0, 1]$, forming  $\boldsymbol{x}_t = \alpha_t \boldsymbol{x}_0 + \beta_t \boldsymbol{x}_1$. We adopt linear schedules $(\alpha_t,\beta_t)=(1-t,t)$ for tabular datasets and the variance-preserving (VP) schedule \cite{song2020score} for all others, following standard practice \cite{choi2022density,chen2025dequantified}.

To thoroughly evaluate model behavior, we consider two categories of benchmarks. (1) Structured and multi-modal datasets: nine synthetic datasets, including  $\mathsf{swissroll}$, $\mathsf{circles}$, $\mathsf{rings}$, $\mathsf{moons}$, $\mathsf{8gaussians}$, $\mathsf{pinwheel}$, $\mathsf{2spirals}$, $\mathsf{checkerboard}$ (first eight following \cite{chen2025dequantified}) and $\mathsf{tree}$ (setup from \cite{bansal2023guiding}). They cover multi-modal, disconnected, discontinuous, and branching geometries. (2)  High-discrepancy real-world tabular datasets: five UCI benchmarks, including  $\mathsf{POWER}$, $\mathsf{GAS}$, $\mathsf{HEPMASS}$, $\mathsf{MINIBOONE}$, and $\mathsf{BSDS300}$ \cite{grathwohl2018ffjord}. These datasets contain complex correlations and are standard for testing DRE scalability.

\subsubsection{Results on Structured and Multi‑modal Datasets}
\begin{figure}[ht]
	\centering
	\includegraphics[width=0.98\linewidth]{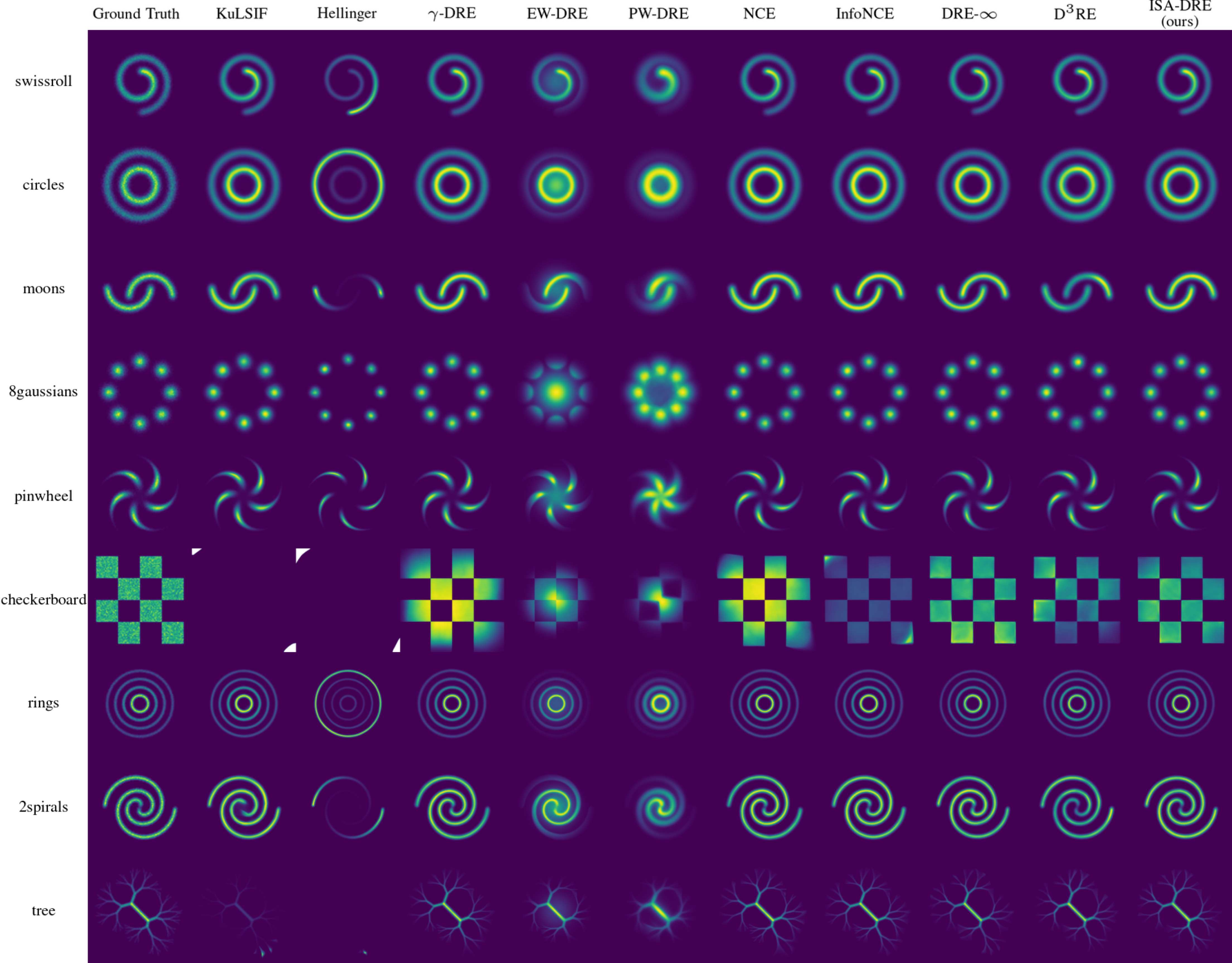}
	\caption{Qualitative density estimates on structured multi-modal data. Baselines (e.g., NCE) blur discontinuities and merge modes. D$^3$RE yields noisy estimates. ISA-DRE preserves structural fidelity and accurately captures density chasms.
	}
	\label{fig:likelihoodtoy2dablationmethods}
\end{figure}

We visualize the estimated densities in \cref{fig:likelihoodtoy2dablationmethods} to qualitatively compare the ability of different frameworks to recover complex data topology. The results reveal a clear trade-off in existing baselines when facing the density-chasm problem. 

Discriminative Neural DRE methods (e.g., NCE, PW-DRE) are stable but heavily over-smooth the densities. They fail to capture sharp transitions, often blurring boundaries or connecting disconnected manifolds. This behavior is most evident on the $\mathsf{checkerboard}$ dataset, where these methods cannot separate the disjoint modes, and on $\mathsf{rings}$, where they incorrectly allocate probability mass to low-density gaps.

Tangent-based score estimators such as D$^3$RE exhibit the opposite pathology. The high variance of the instantaneous tangent target produces noisy and incoherent estimates, especially near support boundaries where the density changes sharply.

ISA-DRE avoids both failure modes. By estimating the stable secant integral, it achieves robustness without sacrificing expressivity. On $\mathsf{checkerboard}$, ISA-DRE is the only method that cleanly reconstructs the grid and resolves the discontinuous density chasm. Likewise, on $\mathsf{2spirals}$ and $\mathsf{8gaussians}$, it preserves fine-grained structure that discriminative and tangent-based baselines either smooth out or distort.

\subsubsection{Results on Real-world Tabular Datasets}
\begin{figure*}[!t]
    \centering
    \includegraphics[width=\linewidth]{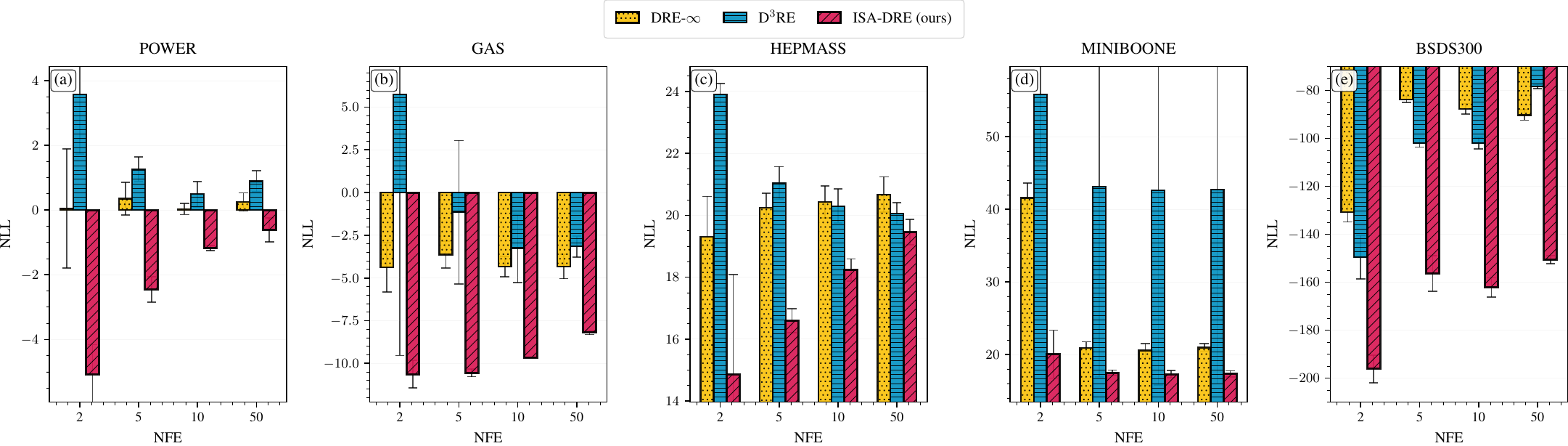}
    \caption{Density estimation (NLL, lower better) on five non-Gaussian tabular datasets (POWER, GAS, HEPMASS, MINIBOONE, BSDS300) across $\text{NFE}\in\{2,5,10,50\}$). Shown: DRE-$\infty$, D$^3$RE, and ISA-DRE (ours).  Error bars: std. over $3$ runs. ISA-DRE consistently achieves the lowest NLL.
}
    \label{fig:density_estimation_tabular_NLL_vs_NFE}
\end{figure*}

We follow the preprocessing and data splits of \cite{grathwohl2018ffjord} and evaluate density estimation using negative log-likelihood (NLL). The quantitative results, summarized in \cref{fig:density_estimation_tabular_NLL_vs_NFE,tab:density_estimation_tabular}, provide strong empirical evidence for the efficacy of our stability-first design.

\begin{table*}[htbp]
\centering
\setlength{\tabcolsep}{1.2mm}
\caption{Test NLL Results on Five Tabular Datasets. Lower is better. Results are reported as $\text{mean} \pm \text{std}$. For score-based methods, $\text{NFE}=50$ results are shown. For ISA-DRE, VI+CIA is selected. N/A indicates abnormal training (e.g., divergence).}
\label{tab:density_estimation_tabular}
\begin{tabular}{clccccc}
\toprule
\textbf{Category} & \textbf{Method} & \textbf{POWER} & \textbf{GAS} & \textbf{HEPMASS} & \textbf{MINIBOONE} & \textbf{BSDS300} \\
\midrule
\multirow{7}{*}{Direct Neural DRE}
& KuLSIF \cite{kanamori2009least} & N/A & N/A & $13.98 \scriptstyle\pm 0.10$ & N/A & $82.88 \scriptstyle\pm 12.72$ \\
& Hellinger \cite{birrell2021variational} & N/A & N/A & N/A & N/A & $-7.70 \scriptstyle\pm 1.12$ \\
& $\gamma$-DRE \cite{nagumo2024density} & $2.91 \scriptstyle\pm 723.16$ & N/A & N/A & N/A & N/A \\
& EW-DRE \cite{zellinger2025binary} & $6.24 \scriptstyle\pm 0.00$ & $9.14 \scriptstyle\pm 0.00$ & $27.51 \scriptstyle\pm 0.00$ & $58.50 \scriptstyle\pm 0.00$ & $55.85 \scriptstyle\pm 0.00$ \\
& PW-DRE \cite{zellinger2025binary} & $6.64 \scriptstyle\pm 0.01$ & $9.13 \scriptstyle\pm 0.00$ & $27.66 \scriptstyle\pm 0.00$ & $58.48 \scriptstyle\pm 0.00$ & $55.85 \scriptstyle\pm 0.00$ \\
& NCE \cite{gutmann2010noise} & $0.03 \scriptstyle\pm 0.03$ & $-6.11 \scriptstyle\pm 0.06$ & $\mathbf{18.43} \scriptstyle\pm 0.03$ & $26.01 \scriptstyle\pm 0.72$ & $3.04 \scriptstyle\pm 0.32$ \\
& InfoNCE \cite{oord2018representation} & $12.57 \scriptstyle\pm 3.05$ & $12.90 \scriptstyle\pm 0.55$ & $32.84 \scriptstyle\pm 0.25$ & $48.93 \scriptstyle\pm 2.10$ & $29.35 \scriptstyle\pm 0.00$ \\
\midrule
\multirow{3}{*}{Score-based DRE}
& DRE-$\infty$ \cite{choi2022density} & $0.25 \scriptstyle\pm 0.28$ & $-4.33 \scriptstyle\pm 0.71$ & $20.67 \scriptstyle\pm 0.57$ & $20.97 \scriptstyle\pm 0.51$ & $-90.24 \scriptstyle\pm 2.14$ \\
& D$^3$RE \cite{chen2025dequantified} & $0.89 \scriptstyle\pm 0.33$ & $-3.16 \scriptstyle\pm 0.62$ & $20.05 \scriptstyle\pm 0.35$ & $42.73 \scriptstyle\pm 26.78$ & $-78.26 \scriptstyle\pm 0.96$ \\
& ISA-DRE (ours) & $\mathbf{-0.61} \scriptstyle\pm 0.37$ & $\mathbf{-8.19} \scriptstyle\pm 0.12$ & $19.46 \scriptstyle\pm 0.41$ & $\mathbf{17.34} \scriptstyle\pm 0.41$ & $\mathbf{-150.54} \scriptstyle\pm 1.71$ \\
\bottomrule
\end{tabular}
\end{table*}

The benefit of the secant formulation is most evident in the low-NFE regime, where high-discrepancy behaviors typically destabilize tangent-based methods. At $\text{NFE}=2$, ISA-DRE achieves large error reductions: from $41.55$ to $20.05$ on $\mathsf{MINIBOONE}$ (a $51.7\%$ improvement) and from $-149.53$ to $-234.21$ on $\mathsf{BSDS300}$ ($56.6\%$ improvement). These results directly reflect the theoretical variance reduction of the secant target, which allows accurate estimation without relying on dense numerical integration to suppress derivative noise.

As compute increases, the VI sampler further enhances efficiency. At $\text{NFE}=50$, our full configuration (VI+CIA) achieves SOTA results on $\mathsf{GAS}$ ($-8.19$) and $\mathsf{BSDS300}$ ($-150.54$), outperforming the strongest baselines by more than $25\%$. Moreover, ISA-DRE typically reaches near-optimal NLL by $\text{NFE}=10$, while baselines require $\text{NFE}=50$ to achieve comparable performance. This efficiency comes from prioritizing low-variance intervals, concentrating learning on the most stable parts of the trajectory and accelerates convergence.

The advantages hold consistently across all datasets and budgets. ISA-DRE achieves the best NLL in $18$ of the $20$ evaluated settings and exhibits smaller variance across runs, particularly on challenging datasets such as $\mathsf{MINIBOONE}$ and $\mathsf{HEPMASS}$. These results demonstrate that the stability gains from the secant objective and VI sampling translate into reliable optimization behavior and improved generalization across diverse tabular distributions.

Crucially, when compared with a broad set of direct neural DRE baselines (\cref{tab:density_estimation_tabular}), ISA-DRE is the only method that produces valid and low-variance estimates across all five tabular datasets. Most direct approaches fail to train stably on at least three benchmarks (reported as N/A), underscoring the difficulty of DRE on high-discrepancy, non-Gaussian tabular data. Among the few workable methods, NCE performs reasonably on $\mathsf{HEPMASS}$ ($18.43$) but collapses on others. For example, $0.03$ on $\mathsf{POWER}$ vs. our $-0.61$, and $-6.11$ on $\mathsf{GAS}$ vs. our $-8.19$. In contrast, ISA-DRE converges reliably on all datasets and achieves the best NLL on four out of five, with markedly lower variance. This demonstrates that its stability-first design yields both robustness and SOTA accuracy in practical DRE tasks.

\subsection{Results for Mutual information Estimation}
\label{sec:mi_estimation}
Mutual information (MI) quantifies the statistical dependence between random variables $\mathbf{x}\sim p(\boldsymbol{x})$ and $\mathbf{y}\sim q(\boldsymbol{y})$, defined as $\text{MI}(\mathbf{x},\mathbf{y}) = \mathbb{E}_{p(\boldsymbol{x},\boldsymbol{y})}\left[\log\frac{p(\boldsymbol{x},\boldsymbol{y})}{p(\boldsymbol{x})q(\boldsymbol{y})}\right]$. Since MI involves the expectation of a log density ratio, its estimation reduces naturally to a DRE problem. We use this task to stress-test our method under challenging geometric and statistical conditions.

\subsubsection{Experimental Setup}
We sample pairs $(\boldsymbol{x}_0, \boldsymbol{x}_1)\sim p_0\times p_1$, and generate interpolated samples $\boldsymbol{x}_t$ using the VP  schedule \cite{song2020score}.
We consider two benchmarks.
First, for topological robustness, we adopt four geometrically pathological distributions from  \cite{czyz2023beyond}: $\mathsf{AsinhMapping}$, $\mathsf{AdditiveNoise}$, $\mathsf{HalfCubeMap}$, and $\mathsf{EdgeSingularGauss}$, each supported on complex nonlinear manifolds.
Second, to probe robustness to the density-chasm problem, we construct high-discrepancy Gaussian pairs with block-diagonal covariances. Each $2\times 2$ block is $\Lambda=\left[\begin{smallmatrix}1 & \rho \\ \rho & 1\end{smallmatrix}\right]$ with $\rho=0.5$. Varying  $d\in\{40, 80, 120, 160 \}$ yields $\text{MI} \geq 20$ nats. The resulting localized dependencies and global sparsity produce ill-conditioned covariances that challenge gradient-based estimators~\cite{choi2022density}.

\begin{figure*}[t]
\centering
\subfloat[Convergence analysis (MSE vs. NFE)\label{fig:different_methods_MSE_vs_NFE}]{%
  \includegraphics[width=0.49\linewidth]{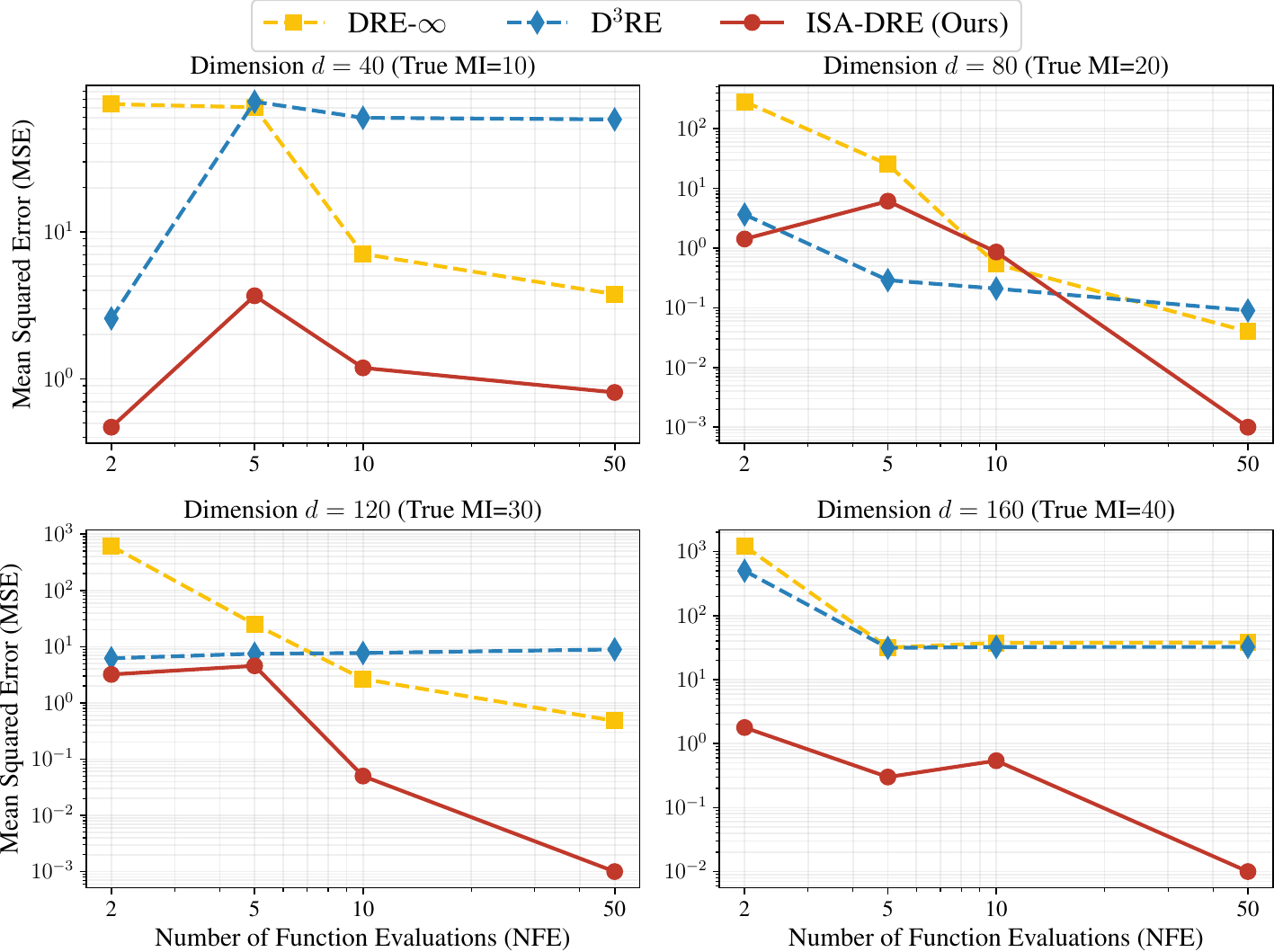}%
}%
\hfill
\subfloat[Robustness at minimal budget ($\text{NFE}=2$)\label{fig:different_methods_MSE_vs_dim}]{%
  \includegraphics[width=0.49\linewidth]{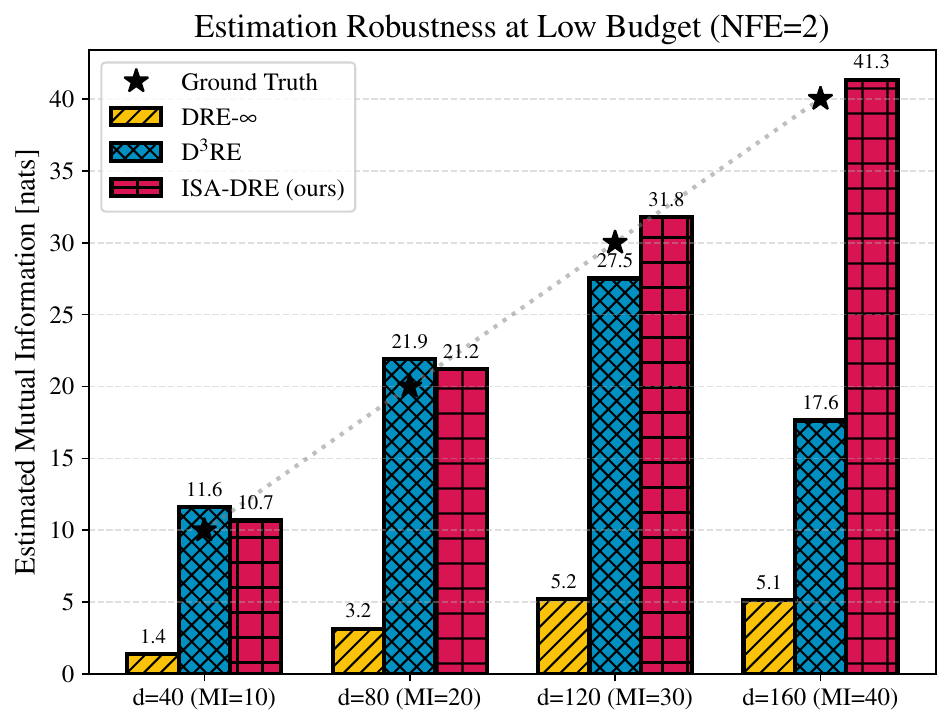}%
}%
\caption{
Robustness of MI estimation on high-discrepancy benchmarks with $d \in \{40,80,120,160\}$ (True MI $\in$ $\{10,20,30,40\}$).  
\textbf{(a)} MSE vs. NFE: DRE-$\infty$ and D$^3$RE saturate or diverge with increasing $d$, while ISA-DRE converges stably to lower error. 
\textbf{(b)} MI estimates at $\text{NFE}=2$ (true MI marked by $\star$). Baselines collapse as the density chasm widens (e.g., at $\text{MI}=40$). ISA-DRE remains close to ground truth.
}
\label{fig:comparison-methods-high-MI-MSE}
\end{figure*}

\subsubsection{Results on Geometrically Pathological Distributions}
\begin{figure}[htbp]
\centering
\subfloat[$\mathsf{AsinhMapping}$\label{fig:comparison-asinh-mapping}]{%
  \includegraphics[width=0.49\linewidth]{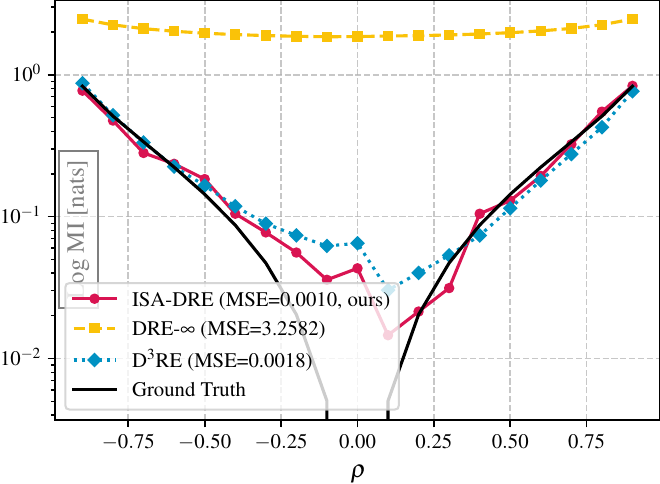}%
}%
\hfill
\subfloat[$\mathsf{AdditiveNoise}$\label{fig:comparison-additive-noise}]{%
  \includegraphics[width=0.49\linewidth]{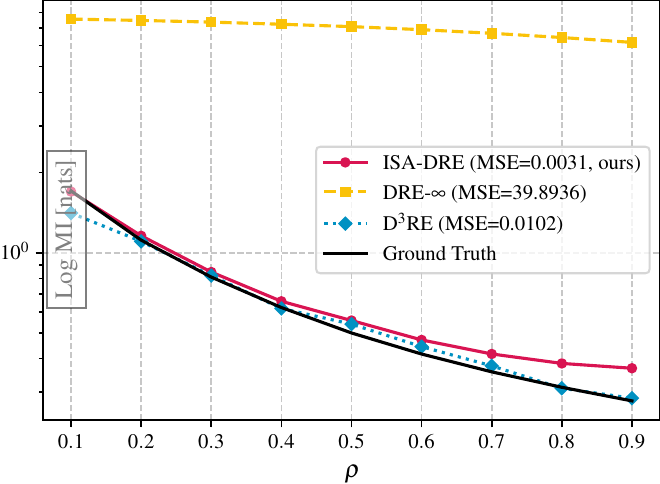}%
}%

\subfloat[$\mathsf{HalfCubeMap}$\label{fig:comparison-half-cube-map}]{%
  \includegraphics[width=0.49\linewidth]{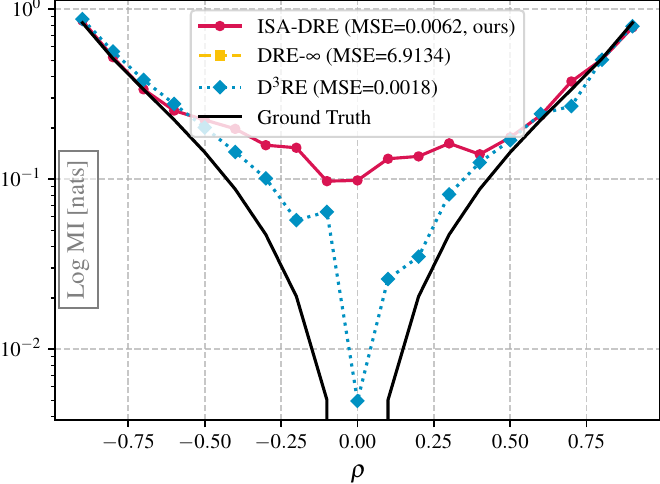}%
}%
\hfill
\subfloat[$\mathsf{EdgeSingularGauss}$\label{fig:comparison-gauss}]{%
  \includegraphics[width=0.49\linewidth]{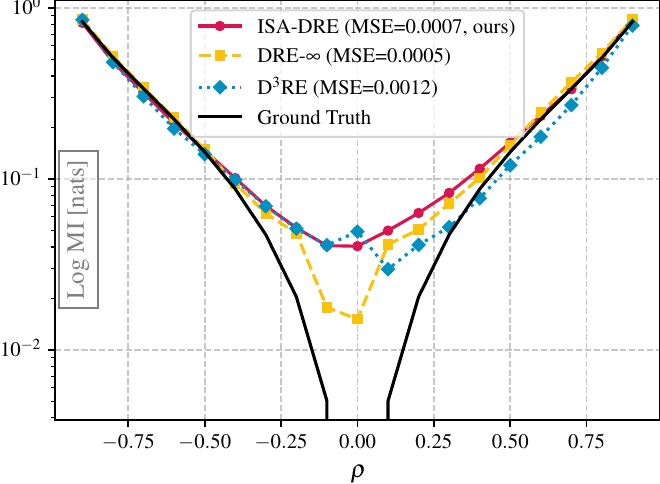}%
}%
\caption{Comparison of ISA-DRE (ours) with DRE-$\infty$ and D$^3$RE on four geometrically pathological distributions. Mean Squared Error (MSE) values for each method are reported. \textit{ISA-DRE achieves the best or comparable results.}
}
\label{fig:comparison-methods-different-noise}
\end{figure}

We verify the method's capability to handle complex manifolds in \cref{fig:comparison-methods-different-noise}. The results strongly support our hypothesis regarding target stability. DRE-$\infty$, which relies on the high-variance tangent target, struggles significantly (e.g., MSEs of $39.89$ on $\mathsf{AdditiveNoise}$), as the pathological geometry exacerbates the variance of instantaneous score estimates. In contrast, ISA-DRE demonstrates superior robustness. On $\mathsf{AsinhMapping}$ and $\mathsf{AdditiveNoise}$, our method achieves the lowest MSE ($0.0010$ and $0.0031$), outperforming baselines by orders of magnitude. This confirms that the secant target effectively smooths out the geometric irregularities that destabilize tangent-based learning.

\subsubsection{Results on High-Discrepancy Distributions}
We further evaluate performance in extreme regimes ($d=160, \text{MI}=40$), which are known to trigger the density-chasm problem \cite{rhodes2020telescoping}. The results, presented in \cref{fig:comparison-methods-high-MI-MSE}, provide the clearest empirical validation of ISA-DRE. As illustrated in \cref{fig:different_methods_MSE_vs_NFE}, while tangent-based baselines (DRE-$\infty$, D$^3$RE) suffer from error saturation or divergence as dimensionality increases, ISA-DRE exhibits consistent monotonic error reduction. This robustness is most pronounced at a minimal budget of $\text{NFE}=2$ (\cref{fig:different_methods_MSE_vs_dim}), where baselines fail catastrophically (MSE $=1215.69$ and $500.04$, respectively) due to the instability of instantaneous tangents in the density chasm. 
In stark contrast, ISA-DRE consistently succeeds where others collapse, maintaining near-perfect estimation (MSE $=0.72$) under the same conditions. Crucially, this result at $\text{NFE}=2$ is a direct consequence of training stability. Our low-variance secant target allows the network to learn a high-fidelity integral approximation  directly, thereby bypassing the need for high-NFE numerical integration to average out estimation noise. This 3-orders-of-magnitude improvement in MSE ($0.72$ for ISA-DRE vs. $1215.69$ for DRE-$\infty$) confirms that ISA-DRE's low-variance framework is more robust to the density-chasm problem.

\subsection{Ablation Study}
\label{section:ablation-study}

\begin{figure}[ht]
    \centering
    \includegraphics[width=1\linewidth]{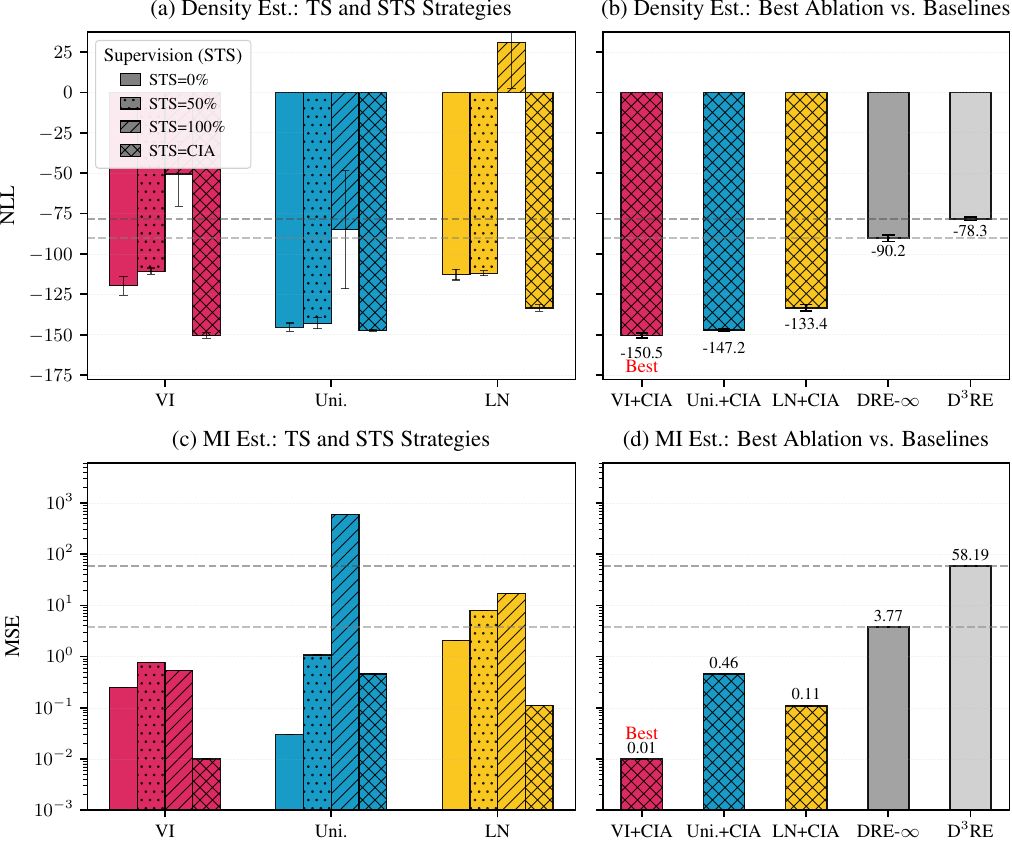}
    \caption{
    Ablation studies on density  ($\mathsf{BSDS300}$, $\text{NFE}=50$) and MI estimation ($\text{MI}=40$, $\text{NFE}=50$).
    \textbf{(a, c)} Varying time samplers (TS: VI, Uniform, LN) and supervision strategies ($\{0, 50, 100\}\%$ STS, CIA); visual encoding: color=TS, hatch=STS.  
    \textbf{(b, d)} Best-performing TS-STS pairs (all using CIA) vs. baselines. VI+CIA achieves lowest NLL ($-150.54$) and MSE ($0.01$). 
    VI and CIA consistently yield best accuracy and stability.
}
\label{fig:ablation_studies_MI_and_density}
\end{figure}

We conduct comprehensive ablation studies on density estimation ($\mathsf{BSDS300}$, $\text{NFE}=50$) and MI estimation ($d=160$, $\text{MI}=40$, $\text{NFE}=50$), evaluating the two core components of ISA-DRE: the \textbf{\textit{VI time sampler}} and the \textbf{\textit{CIA supervision}} strategy. 
Across all experiments (see \cref{fig:ablation_studies_MI_and_density}), two patterns consistently emerge: (i) VI sampling strongly improves optimization stability and final accuracy, and (ii) CIA offers the best trade-off between stability and expressivity.
To better understand these behaviors, we analyze each component separately.

\subsubsection{Impact of Variance-based Sampling}
\label{sec:ablation_vi_sampler}
We compare VI sampling with Uniform and LN across both density estimation and MI estimation. VI biases interval selection toward low score-variance regions, which improves stability by avoiding high-variance time steps during early training. This systematically enhances convergence across settings. In density estimation, VI achieves the best NLL under every supervision configuration. At NFE=50 on $\mathsf{BSDS300}$, VI combined with CIA reaches NLL $-150.54$, improving Uniform+CIA by $3.33$ nats and LN+CIA by $17.13$ nats. Consistent trends appear in high-discrepancy  MI estimation. Under the challenging MI=40 and $d=160$ setting, VI+CIA attains MSE $0.01$, whereas Uniform+CIA and LN+CIA yield $0.46$ and $0.11$. Even at extreme low-compute settings such as $\text{NFE}=2$, VI remains stable while Uniform often fails to converge and LN deteriorates significantly. These results confirm that allocating sampling mass toward low-variance regions is essential for navigating the density-chasm structure, improving both numerical stability and sample efficiency.

\subsubsection{Impact of Supervision Strategy}

We analyze the stability of different supervision targets: tangent-only ($100\%$ STS), secant-only ($0\%$ STS), and our proposed CIA supervision. The comparison yields three key insights that empirically validate our theoretical framework: (i)
\textit{Tangent-only supervision is inherently unstable.}
Consistent with our critique of prior art, $100\%$ STS frequently leads to divergence. For example, at $d=120$, $\text{NFE}=2$, the MSE explodes to $2804.96$. This empirically proves the high-variance nature of the instantaneous tangent target, which cannot be reliably learned in high-discrepancy regimes.
(ii) \textit{Secant-only supervision exhibits a statistical-optimization trade-off.}
Using only the secant target ($0\%$ STS) greatly stabilizes training in well-conditioned settings: in the $d=160$ Gaussian setting, MSE drops to $0.72$, a $77\times$ improvement over the tangent baseline. This validates \cref{thm:variance-reduction-of-secant} that the secant is a lower-variance target.
Yet, the same approach fails to converge on the topologically complex $\mathsf{AdditiveNoise}$ manifold (\cref{fig:ablation-IA-epoch}). This paradox arises because while the secant reduces statistical variance, its optimization via SAI introduces a feedback loop through the derivative correction (\cref{sec:cia}). On complex manifolds, large initial intervals amplify errors, causing ``bootstrap divergence'' despite the target’s theoretical advantages.
(iii) \textit{CIA resolves this trade-off to achieve the optimal balance.}
Our CIA supervision effectively bridges this gap. By enforcing small intervals ($|t-l| \to 0$) during the warm-up, CIA ensures algorithmic stability akin to tangent learning. As training progresses, it gradually expands the interval to leverage the statistical efficiency of the secant. Consequently, at $d=160, \text{NFE}=50$, CIA achieves near-perfect estimation ($\text{MSE}=0.01$), outperforming the static $0\%$ baseline. These results demonstrate that CIA is not merely a heuristic, but a structural necessity that reaches an effective balance between stability and accuracy.

\section{Conclusion and Future Work}
We proposed ISA-DRE, a framework aimed at a core limitation of modern score-based DRE: the inherently high-variance learning target used in tangent-based methods. Prior work largely addressed its downstream consequence (high inference cost), while the underlying issue remained unsolved. By analyzing the learning dynamics themselves, we identify the instability as a structural property of predicting the instantaneous tangent. 
ISA-DRE replaces this fragile target with a more stable and theoretically grounded quantity. Rather than learning the tangent, it learns its interval integral, the secant. As shown in \cref{thm:variance-reduction-of-secant}, the secant enjoys strictly lower variance and smoother behavior, making it a better match for neural approximation.
To make secant learning practicable, we introduced two key components. The Secant Alignment Identity (SAI) provides a self-consistency relation that converts the secant into a supervised objective.
The Contraction Interval Annealing (CIA) mechanism prevents the bootstrap instability induced by SAI and ensures convergence even when the underlying path is difficult to learn.
The empirical results align closely with the theoretical predictions. In regimes with large density discrepancies, tangent-based estimators exhibit the expected failure modes tied to their variance. ISA-DRE remains stable, achieves higher accuracy, and reduces the number of function evaluations needed for inference, demonstrating that variance reduction at the modeling level leads directly to computational efficiency.
Several open questions remain. ISA-DRE, like prior DRE methods, can be sensitive to the choice of interpolant, and understanding how to design or learn optimal paths is an important direction for future work.




\appendix

\subsection{Proof of \cref{thm:variance-reduction-of-secant}}
\label{proof:variance-reduction-of-secant}


\begin{proof}
	To establish the variance inequality, we introduce an auxiliary random variable $\eta$ whose conditional distribution given $l$ and $t$ is uniform on $[l, t]$, denoted as $\eta \mid (l, t) \sim \mathcal{U}[l, t]$. Define the composite random variable $Z = s^{(t)}(\boldsymbol{x}, \eta)$.

    We now apply the law of total variance to $S$, conditioning on the $\sigma$-algebra generated by $(l, t)$. The decomposition yields:
	\begin{equation}  \label{eq:variance-decomp-of-tangent}
		\begin{aligned}
			\mathrm{Var}_{p(\tau)}(S) =& \mathbb{E}_{p(l, t)} \left[ \mathrm{Var}_{p(\tau)}\left( Z \mid l, t \right) \right] \\
			&+ \mathrm{Var}_{p(l, t)} \left( \mathbb{E}_{p(\tau)} \left[ Z \mid l, t \right] \right),
		\end{aligned}
	\end{equation}
	where all conditional expectations and variances are computed with respect to the uniform measure $\mathcal{U}[l, t]$ on $\eta$. 

    The second term on the right hand side of \cref{eq:variance-decomp-of-tangent} is equal to the variance of the secant function:
	\begin{equation}
		\begin{aligned}
			\mathbb{E}_{p(\tau)} \left[ Z \mid l, t \right] &= \mathbb{E}_{\eta \sim p(\tau)} \left[ s^{(t)}(\boldsymbol{x}, \eta) \mid l, t \right] \\
			&= \frac{1}{t-l} \int_l^t s^{(t)}(\boldsymbol{x}, \tau)  \mathrm{d}\tau \\
			&= u(\boldsymbol{x}, l, t) = U.
		\end{aligned}
	\end{equation}

    Substituting this identity into \cref{eq:variance-decomp-of-tangent}:
	\begin{equation}
		\begin{aligned}
			\mathrm{Var}_{p(\tau)}(S) = \mathbb{E}_{p(l, t)} \left[ \mathrm{Var}_{p(\tau)}\left( Z \mid l, t \right) \right] + \mathrm{Var}_{p(l, t)}(U).
		\end{aligned}
	\end{equation}

    For the first term on the right hand side of \cref{eq:variance-decomp-of-tangent}, the conditional variance term is non-negative:
	\begin{equation}
		\begin{aligned}
			\mathrm{Var}_{p(\tau)}\left( Z \mid l, t \right) &= \mathrm{Var}_{\eta\sim p(\tau)}\left( s^{(t)}(\boldsymbol{x}, \eta) \mid l, t \right) \\
			&= \mathbb{E}_{p(\tau)} \left[ \left| s^{(t)}(\boldsymbol{x}, \eta) - U \right|^2 \mid l, t \right] \geq 0.
		\end{aligned}
	\end{equation}
	which implies:
	\begin{equation}
		\mathrm{Var}_{p(\tau)}(S) \geq \mathrm{Var}_{p(l, t)}(U).
	\end{equation}


    Equality holds if and only if for $p$-almost every pair $(l, t)$, i.e., $\mathrm{Var}_{\eta\sim\mathcal{U}[l, t]}\left( s^{(t)}(\boldsymbol{x}, \eta) \mid l, t \right) = 0$.
    Since the intervals $[l, t]$ densely cover $[0,1]$ as $(l,t)$ ranges over the support of $p(l,t)$, the requirement that $s^{(t)}(\boldsymbol{x},\cdot)$ be constant Lebesgue-a.e. on each $[l,t]$ implies $s^{(t)}(\boldsymbol{x},\tau)$ is constant for $p$-almost every $\tau \in [0,1]$.

\end{proof}

\subsection{Proof of \cref{prop:smoothness-of-secant}}
\label{proof:smoothness-of-secant}

\begin{proof}
	Fix $\boldsymbol{x}$ and $l$, and consider $t_0, t_1\in (l,1]$. Define the auxiliary function $g(t) = \int_l^t s^{(t)}(\boldsymbol{x}, \tau)  \mathrm{d}\tau$, so that the secant function can be expressed as $u(\boldsymbol{x}, l, t) = \frac{1}{t - l}g(t)$.
	
	Then, the derivative of $u$ with respect to $t$ is given by,
	\begin{equation} \label{eq:derivative-of-secant-function}
		\frac{\mathrm{d} u}{\mathrm{d} t} = \frac{ g^{\prime}(t) (t - l) - g(t) }{ (t - l)^2 }.
	\end{equation}
	
	By the Fundamental Theorem of Calculus, $g^{\prime}(t) = s^{(t)}(\boldsymbol{x}, t)$, thus the absolute value of numerator, i.e., $g^{\prime}(t) (t - l) - g(t)$, satisfies:
	\begin{equation}
		\begin{aligned}
			&\quad\left|g^{\prime}(t) (t - l) - g(t)\right|\\
			&= \left| s^{(t)}(\boldsymbol{x}, t) (t - l) - \int_l^t s^{(t)}(\boldsymbol{x}, \tau)  \mathrm{d}\tau \right| \\
			&= \left|\int_l^t \left[ s^{(t)}(\boldsymbol{x}, t) - s^{(t)}(\boldsymbol{x}, \tau) \right] \mathrm{d}\tau \right| \\
			&\le \int_l^t \left| s^{(t)}(\boldsymbol{x}, t) - s^{(t)}(\boldsymbol{x}, \tau) \right| \mathrm{d}\tau\\
			&\le \int_l^t \lambda |t - \tau| \mathrm{d}\tau \quad (\star) \\
			&=\lambda \left( \frac{1}{2}t^2 - tl + \frac{1}{2}l^2 \right)=\frac{\lambda}{2} (t - l)^2,
		\end{aligned}
	\end{equation}
	where $(\star) $ holds because the score function $s^{(t)}(\boldsymbol{x}, \tau)$ is $\lambda$-Lipschitz continuous in $\tau$, which leads to $| s^{(t)}(\boldsymbol{x}, t) - s^{(t)}(\boldsymbol{x}, \tau) | \leq \lambda |t - \tau|, \forall \tau \in [l, t]$.
	
	Substituting this bound into \cref{eq:derivative-of-secant-function}:
	\begin{equation}
		\left| \frac{\mathrm{d} u}{\mathrm{d} t} \right| \leq \frac{ \frac{\lambda}{2} (t - l)^2 }{ (t - l)^2 } = \frac{\lambda}{2}.
	\end{equation}
	
	By the Mean Value Theorem, for any $t_0, t_1\in (l,1]$:
	\begin{equation}
		\begin{aligned}
			|u(\boldsymbol{x}, l, t_1) - u(\boldsymbol{x}, l, t_0)| &\leq \sup_{\xi \in [t_0,t_1]} \left| \frac{\partial u}{\partial t}(\xi) \right| \cdot |t_1 - t_0|\\
			&\leq \frac{\lambda}{2} |t_1 - t_0|,
		\end{aligned}
	\end{equation}
	which completes the proof.
\end{proof}

\subsection{Proof of \cref{proposition:secant-tangent-consistency}}
\label{proof:secant-tangent-consistency}
\begin{proof}
	The proposition is a biconditional statement, which requires us to prove two directions.
	
	First, we prove the forward direction ($\Rightarrow$). 
	
	We assume that $u_{\boldsymbol{\theta}^\star}(\boldsymbol{x}, l, t)$ is identical to the true secant function and demonstrate that it must satisfy both the boundary and consistency conditions. Assume
	\begin{equation}
		u_{\boldsymbol{\theta}^\star}(\boldsymbol{x}, l, t) = \frac{1}{t-l} \int_l^t s^{(t)}(\boldsymbol{x}, \tau) \mathrm{d}\tau.
	\end{equation}
	We begin by verifying the boundary condition. The limit of $u_{\boldsymbol{\theta}^{\star}}(\boldsymbol{x}, t_0, t)$ as $t \to t_0$ takes the indeterminate form $\frac{0}{0}$. We can therefore apply L'Hôpital's Rule:
	\begin{equation}
		\begin{aligned}
			\lim_{t \to t_0} u_{\boldsymbol{\theta}^{\star}}(\boldsymbol{x}, t_0, t) &= \lim_{t \to t_0} \frac{\int_{t_0}^t s^{(t)}(\boldsymbol{x}, \tau)  \mathrm{d}\tau}{t-t_0} \\
			&= \lim_{t \to t_0} \frac{\frac{\partial}{\partial t}\left(\int_{t_0}^t s^{(t)}(\boldsymbol{x}, \tau)  \mathrm{d}\tau\right)}{\frac{\partial}{\partial t}(t-t_0)} \\
			&=\lim_{t \to t_0} \frac{s^{(t)}(\boldsymbol{x}, t)}{1} = s^{(t)}(\boldsymbol{x}, t_0).
		\end{aligned}
	\end{equation}
	This confirms that the boundary condition is satisfied.
	
	Next, we verify the consistency condition. We must show that the expression $u_{\boldsymbol{\theta}^{\star}}(\boldsymbol{x}, l, t) + (t-l) \frac{\mathrm{d}}{\mathrm{d} t} u_{\boldsymbol{\theta}^{\star}}(\boldsymbol{x}, l, t)$ simplifies to $s^{(t)}(\boldsymbol{x}, t)$. To do this, we first compute the partial derivative $\frac{\mathrm{d} u_{\boldsymbol{\theta}^{\star}}}{\mathrm{d} t}$ using the product rule for differentiation on the expression $u_{\boldsymbol{\theta}^{\star}}(\boldsymbol{x}, l, t) = \frac{1}{t-l} \int_l^t s^{(t)}(\boldsymbol{x}, \tau)  \mathrm{d}\tau$:
	\begin{equation}
		\begin{aligned}
			&\frac{\mathrm{d}}{\mathrm{d} t} u_{\boldsymbol{\theta}^{\star}}(\boldsymbol{x}, l, t)\\ =& \frac{\mathrm{d}(t-l)^{-1}}{\mathrm{d} t}\int_l^t s^{(t)}(\boldsymbol{x},\tau)  \mathrm{d}\tau + \frac{1}{t-l}  \frac{\mathrm{d}}{\mathrm{d} t}\int_l^t s^{(t)}(\boldsymbol{x},\tau) \mathrm{d}\tau \\
			=& \frac{-1}{(t-l)^2} \int_l^t s^{(t)}(\boldsymbol{x},\tau) \mathrm{d}\tau + \frac{1}{t-l}  s^{(t)}(\boldsymbol{x}, t) \\
			=& \frac{-1}{(t-l)^2} \int_l^t s^{(t)}(\boldsymbol{x}, \tau) \mathrm{d}\tau + \frac{s^{(t)}(\boldsymbol{x}, t)}{t-l}.
		\end{aligned}
	\end{equation}
	
	Now, substituting this derivative back into the consistency condition expression yields:
	\begin{equation}
		\begin{aligned}
			&u_{\boldsymbol{\theta}^{\star}}(\boldsymbol{x}, l, t) + (t-l) \frac{\mathrm{d}}{\mathrm{d} t} u_{\boldsymbol{\theta}^{\star}}(\boldsymbol{x}, l, t) \\ 
			=& \frac{1}{t-l} \int_l^t s^{(t)}(\boldsymbol{x}, \tau) \mathrm{d}\tau - \frac{1}{t-l} \int_l^t s^{(t)}(\boldsymbol{x}, \tau) \mathrm{d}\tau + s^{(t)}(\boldsymbol{x}, t) \\
			=& s^{(t)}(\boldsymbol{x}, t).
		\end{aligned}
	\end{equation}
	
	The consistency condition is therefore also satisfied. This completes the proof of the forward direction.
	
	Finnaly, we now prove the reverse direction ($\Leftarrow$). 
	
	We assume that a function $u_{\boldsymbol{\theta}^{\star}}$ satisfies both the boundary and consistency conditions. Our objective is to prove that $u_{\boldsymbol{\theta}^{\star}}$ must be uniquely determined and equal to the true secant function. The consistency condition itself is a differential equation that governs the evolution of $u_{\boldsymbol{\theta}^{\star}}$ with respect to time $t$. For $t \neq l$, we can rearrange the equation into the standard form of a first-order linear ordinary differential equation:
	\begin{equation}
		\frac{\mathrm{d}}{\mathrm{d} t} u_{\boldsymbol{\theta}^{\star}}(\boldsymbol{x}, l, t) + \frac{1}{t-l} u_{\boldsymbol{\theta}^{\star}}(\boldsymbol{x}, l, t) = \frac{1}{t-l} s^{(t)}(\boldsymbol{x}, t).
	\end{equation}
	
	This type of differential equation can be solved using the method of integrating factors. The integrating factor, denoted $I(t)$, is given by:
	\begin{equation}
		I(t) = \exp\left(\int \frac{1}{t-l}  \mathrm{d}t\right)= \exp(\ln|t-l|) = |t-l|.
	\end{equation}
	
	Without loss of generality, let us consider the case where $t > l$, so the integrating factor is $(t-l)$. Multiplying the standard-form ODE by this factor yields:
	\begin{equation}
		\begin{aligned}
			(t-l) \frac{\mathrm{d} u_{\boldsymbol{\theta}^{\star}}(\boldsymbol{x}, l, t)}{\mathrm{d} t} + u_{\boldsymbol{\theta}^{\star}}(\boldsymbol{x}, l, t) &= s^{(t)}(\boldsymbol{x}, t) \\
			\Rightarrow\frac{\mathrm{d}}{\mathrm{d} t}\left[ (t-l) u_{\boldsymbol{\theta}^{\star}}(\boldsymbol{x}, l, t) \right] &= s^{(t)}(\boldsymbol{x}, t).
		\end{aligned}
	\end{equation}
	
	We can now integrate both sides with respect to $\tau$ from the initial point $l$ to a generic endpoint $t^\prime$:
	\begin{equation}
		\int_l^{t^\prime} \frac{\mathrm{d}}{\mathrm{d} \tau} \left[ (\tau-l) u_{\boldsymbol{\theta}^{\star}}(\boldsymbol{x}, l, \tau) \right] \mathrm{d}\tau = \int_l^{t^\prime} s^{(t)}(\boldsymbol{x}, \tau) \mathrm{d}\tau.
	\end{equation}
	
	Applying the Fundamental Theorem of Calculus to the left-hand side gives:
	\begin{equation}
		\left[ (\tau-l) u_{\boldsymbol{\theta}^{\star}}(\boldsymbol{x}, l, \tau) \right]_{\tau=l}^{\tau=t^\prime} = \int_l^{t^\prime} s^{(t)}(\boldsymbol{x}, \tau)  \mathrm{d}\tau.
	\end{equation}
	
	Evaluating the expression at the bounds, we have:
	\begin{equation}
		\begin{aligned}
			& \int_l^{t^\prime} s^{(t)}(\boldsymbol{x}, \tau) \mathrm{d}\tau \\
			=& (t^\prime-l) u_{\boldsymbol{\theta}^{\star}}(\boldsymbol{x}, l, t^\prime) - \lim_{\tau \to l^+} (\tau-l) u_{\boldsymbol{\theta}^{\star}}(\boldsymbol{x}, l, \tau) \\
			=& (t^\prime-l) u_{\boldsymbol{\theta}^{\star}}(\boldsymbol{x}, l, t^\prime) - \lim_{\tau \to l^+} (\tau-l) \cdot \lim_{\tau \to l^+} u_{\boldsymbol{\theta}^{\star}}(\boldsymbol{x}, l, \tau) \\
			= & (t^\prime-l) u_{\boldsymbol{\theta}^{\star}}(\boldsymbol{x}, l, t^\prime) - 0 \cdot s^{(t)}(\boldsymbol{x}, l) \quad (\star)  \\
			=& (t^\prime-l) u_{\boldsymbol{\theta}^{\star}}(\boldsymbol{x}, l, t^\prime).
		\end{aligned}
	\end{equation}
	Here, the equality $(\star)$ holds according to the boundary condition. We know that $\lim_{\tau \to l^+} u_{\boldsymbol{\theta}^{\star}}(\boldsymbol{x}, l, \tau)$ converges to the finite value $s^{(t)}(\boldsymbol{x}, l)$. 
	
	The limit term vanishes, leaving us with a direct algebraic relationship $(t^\prime-l) u_{\boldsymbol{\theta}^{\star}}(\boldsymbol{x}, l, t^\prime) = \int_l^{t^\prime} s^{(t)}(\boldsymbol{x}, \tau)  \mathrm{d}\tau$.
	For any $t^\prime \neq l$, we can divide by $(t^\prime-l)$ to solve for $u_{\boldsymbol{\theta}^{\star}}(\boldsymbol{x}, l, t^\prime)$:
	\begin{equation}
		u_{\boldsymbol{\theta}^{\star}}(\boldsymbol{x}, l, t^\prime) = \frac{1}{t^\prime-l} \int_l^{t^\prime} s^{(t)}(\boldsymbol{x}, \tau) \mathrm{d}\tau.
	\end{equation}
	This exactly matches the definition of the true secant function. By the Picard–Lindelöf theorem, the linear IVP admits a unique solution. Hence any  $u_{\boldsymbol{\theta}^\star}$ satisfying the consistency and boundary conditions must coincide with the true secant function, completing the reverse direction.
\end{proof}

\bibliographystyle{IEEEtran} 
\bibliography{IEEEabrv,references}

\vfill

\end{document}